\documentclass[10pt,twocolumn,letterpaper]{article}

\usepackage{iccv}
\usepackage{times}
\usepackage{epsfig}
\usepackage{graphicx}
\usepackage{amsmath}
\usepackage{amssymb}
\usepackage{titling}

\usepackage[breaklinks=true,bookmarks=false]{hyperref}

\iccvfinalcopy 

\def\iccvPaperID{****} 

\ificcvfinal\fi

\begin{document}

\title{Sampling-free Epistemic Uncertainty Estimation Using Approximated Variance Propagation}

\date{}

\author{Janis Postels\\
Technical University Munich\\
{\tt\small janis.postels@tum.de}
\and
Francesco Ferroni\\
Autonomous Intelligent Driving GmbH\\
{\tt\small francesco.ferroni@aid-driving.eu}
\and
Huseyin Coskun\\
Technical University Munich\\
{\tt\small huseyin.coskun@tum.de}
\and
Nassir Navab\\
Technical University Munich\\
{\tt\small nassir.navab@tum.de}
\and
Federico Tombari\\
Technical University Munich\\
Google\\
{\tt\small tombari@in.tum.de}
}

\maketitle
\ificcvfinal\thispagestyle{empty}\fi

\begin{abstract}
   We present a sampling-free approach for computing the epistemic uncertainty of a neural network. Epistemic uncertainty is an important quantity for the deployment of deep neural networks in safety-critical applications, since it represents how much one can trust predictions on new data. Recently promising works were proposed using noise injection combined with Monte-Carlo sampling at inference time to estimate this quantity (e.g. Monte-Carlo dropout). Our main contribution is an approximation of the epistemic uncertainty estimated by these methods that does not require sampling, thus notably reducing the computational overhead. We apply our approach to large-scale visual tasks (\ie, semantic segmentation and depth regression) to demonstrate the advantages of our method compared to sampling-based approaches in terms of quality of the uncertainty estimates as well as of computational overhead. 
\end{abstract}

\section{Introduction}

\begin{figure}[h]
\begin{tabular}{c@{\hskip 3pt}c@{\hskip 3pt}c}
    \textbf{Image} & \textbf{Prediction} & \textbf{Ours: 0.14s} \\
    \includegraphics[width=25mm]{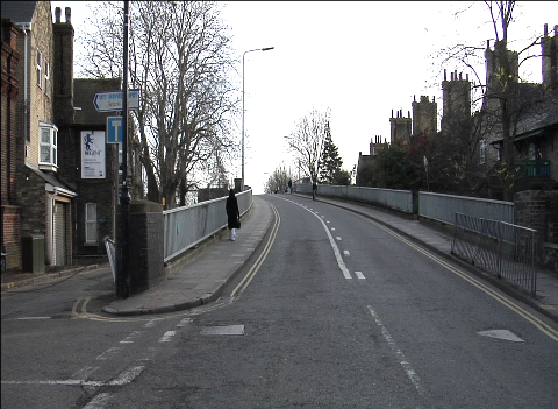}& 
    \includegraphics[width=25mm]{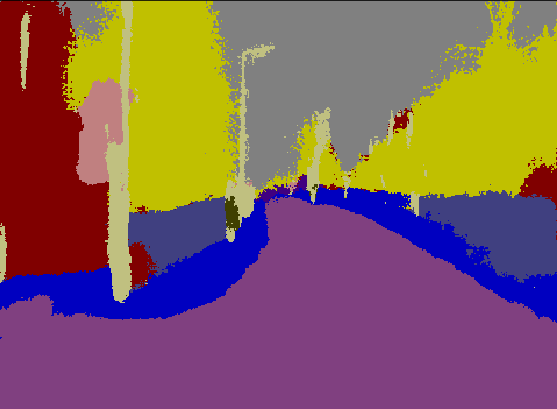}& 
    \includegraphics[width=25mm]{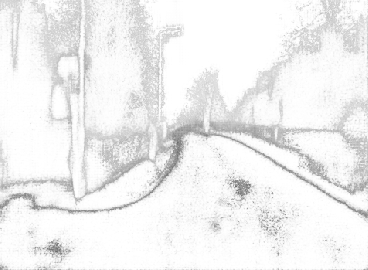}\\
  \includegraphics[width=25mm]{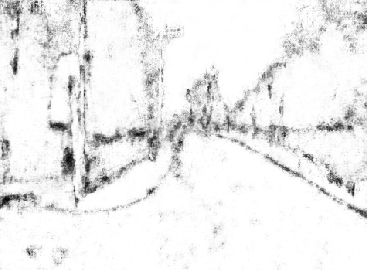} &   \includegraphics[width=25mm]{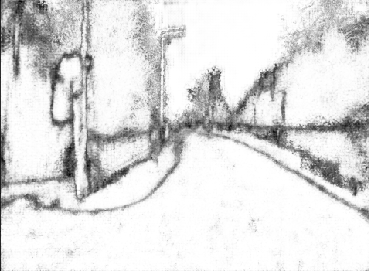} &  \includegraphics[width=25mm]{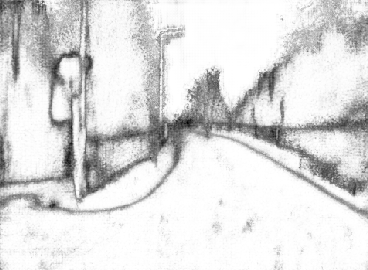}\\
  \textbf{0.13s} & \textbf{0.54s} & \textbf{2.39s} \\
  \textbf{2 samples} & \textbf{10 samples} & \textbf{50 samples} \\
  \multicolumn{3}{c}{\textbf{Monte-Carlo Dropout \cite{gal2016dropout}}}
\end{tabular}
\caption{Qualitative results of uncertainty estimation for Bayesian SegNet (white: small, black: large). Upper: Original image (left), prediction using Monte-Carlo (MC) sampling (middle) and proposed uncertainty prediction (right, with average runtime per prediction). Lower: Uncertainty using MC dropout \cite{gal2016dropout} using 2 (left), 10 (middle) and 50 (right) samples, each with its average runtime per prediction. We cache results prior to the first dropout layer to optimize performance of MC dropout.}\label{teaser_classArchitecure_uncertainty}
\end{figure}

Quantifying the uncertainty associated with the prediction of neural networks is a prerequisite for their deployment and use in safety-critical applications. Whether used to detect road users and make driving decisions in an autonomous vehicle, or in a medical setting within a surgical robot, neural networks must be able not just to predict accurately, but also to quantify how certain they are regarding such predictions. Moreover, it is important that such uncertainty is provided during inference in real-time, so that the uncertainty can be exploited by the real-time safety-critical system.

One can estimate two types of uncertainty of a machine learning model \cite{gal2016uncertainty}: aleatoric and epistemic. 
Aleatoric uncertainty is inherent to the data itself, e.g. uncertainty resulting from noisy sensors. This type of uncertainty can be incorporated into the deep model itself by applying e.g. mixture density networks \cite{bishop1994mixture}.
Epistemic uncertainty is the uncertainty in the chosen model parameters. In order to detect situations which are unfamiliar for a given machine learning model, and consequently quantify how much one can trust predictions on the given data, one has to determine the latter of the two types of uncertainty.

Recently many approaches have been proposed that make it possible to estimate epistemic uncertainty for large scale neural network architectures\cite{gal2016dropout, teye2018bayesian, lakshminarayanan2017simple, osband2016deep}. A promising research direction is the use of noise injection \cite{gal2016dropout, teye2018bayesian} via, \eg, stochastic regularization techniques. The underlying idea is to train a neural network while injecting noise at certain layers. During training, the network learns how to compensate the noise on the training data distribution, and thus minimize the variance of the prediction. At inference time, one can then use the variance in the prediction generated by different noise samples as an epistemic uncertainty estimate, since the neural network only learned to compensate the noise on the training data distribution. 

Unfortunately, while these methods are conceptually simple and have been successful in delivering a measure of epistemic uncertainty for neural networks (even large architectures, e.g. \cite{Kendall2015BayesianSM, gal2017deep, bhattacharyya2018long}), they rely on Monte-Carlo (MC) sampling at inference time in order to determine the variance of the prediction as an uncertainty estimate. This means that computation time scales linearly with the number of samples, and therefore, can become  prohibitively expensive for performance-critical or compute-limited applications, such as autonomous vehicles, robots and mobile devices. In such cases, obtaining epistemic uncertainty as part of a model prediction can be a functional safety requirement, but the need to perform real-time inference from sensor data makes MC dropout a difficult proposition.

In this work, we side-step these issues and produce epistemic uncertainty estimates of a neural network's prediction which are at the same time accurate and computationally inexpensive. Our contributions are specifically:
\begin{itemize}
  \item We propose a sampling-free approach to generally approximate uncertainty estimates of methods that rely on noise injection at training time.
  \item We simplify this approach specifically for the common case of convolutional neural networks combined with rectified linear units as activation function.
\end{itemize}

In the following sections we will firstly outline relevant work, secondly present our sampling-free framework and finally show experimental results. Specifically, we compare the quality of our approximation to Bayesian SegNet \cite{Kendall2015BayesianSM} on CamVid dataset \cite{brostow2009semantic}, and show the ability of our approximation to detect out-of-distribution samples by training Bayesian SegNet while withholding certain classes at training time. We further apply our approximation in a common regression task for computer vision, \ie monocular depth estimation \cite{godard2017unsupervised}.

\section{Related Work}

Recently there has been a wealth of proposals regarding epistemic uncertainty estimation for large-scale neural networks \cite{blundell2015weight, hernandez2015probabilistic, osband2016deep, lakshminarayanan2017simple, gal2016dropout,teye2018bayesian, zhang2018noisy, mozejko2018inhibited, rupprecht2017learning}.
The common goal is to approximate the full posterior distribution of the parameters of a neural network. 

Some works aim to directly learn the parameters of a family of distributions within back-propagation \cite{zhang2018noisy, blundell2015weight, hernandez2015probabilistic}. Another line of research approximates the posterior by training ensembles of neural networks via random changes in the training setup \cite{osband2016deep, lakshminarayanan2017simple} estimating the target distribution by an ensemble of sample distributions \cite{efron1994introduction}.

A different research avenue utilizes stochastic regularization methods to estimate epistemic uncertainty at inference time \cite{gal2016dropout, teye2018bayesian, zhang2018noisy}. The most prominent example is MC dropout \cite{gal2016dropout} - train a dropout regularized neural network, and then, at inference time, keep dropout turned on to estimate the epistemic uncertainty via the variance of the prediction. These approaches have gained popularity due to the simplicity with which they integrate into the current training methodology. Consequently they have been applied to a variety of tasks \cite{gal2017deep, bhattacharyya2018long, Kendall2015BayesianSM, feng2018towards, kampffmeyer2016semantic}. Despite their achievements, they still suffer from large computational overhead at inference time due to sampling, which makes them prohibitively expensive in applications that demand real-time inference from large neural networks. 

\cite{huang2018efficient} optimizes the application of MC dropout on videos. Therefore the authors treat images which are close in time as constant, and thus samples of the same scene. Consequently each image only has to be processed once while performing approximate MC sampling.

Sampling-free estimation of epistemic uncertainty has been only partially covered in literature. \cite{choi2018uncertainty} incorporates sampling-free epistemic uncertainty into the framework of mixture density networks which, following \cite{le2018uncertainty}, suffers from non-convergence for high dimensional problems. Natural Parameter Networks \cite{wang2016natural, hwang2018sampling} can be considered related to our work. Instead of processing point estimates through a neural network, the authors adjust the transformations at each layer to propagate the natural parameters of a pre-defined distribution, e.g. mean and standard deviation of a Gaussian. Notably, these networks are limited in their choice of transformations as they are constrained to preserve the shape of the underlying distribution, and consequently do not generalize naturally to arbitrary large scale architectures. Thus we will not compare against this work since we are interested in approaches applicable to arbitrary neural networks without altering the training process.


\section{Method}

\begin{figure}[h]
    \centering
    \includegraphics[width=0.30\textwidth]{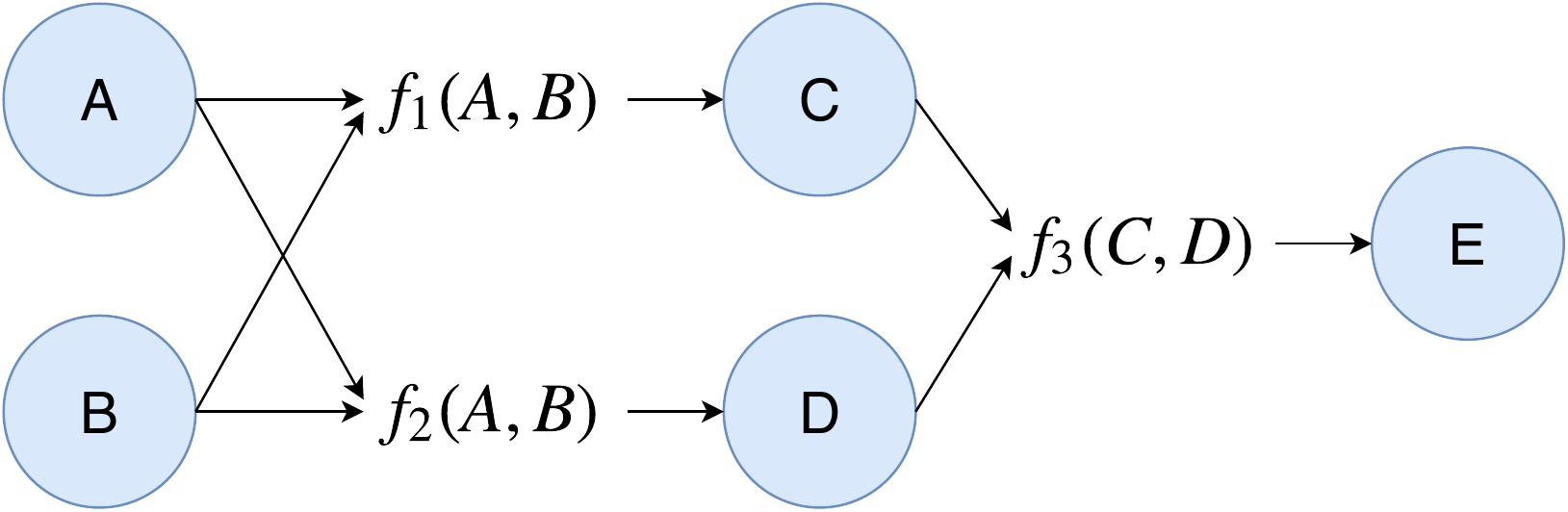}
    \caption{Simple computational graph for illustrating error propagation.}
    \label{fig:simpleerrorpropagation}
\end{figure}

Our goal is to estimate the epistemic uncertainty of a neural network trained with injected noise at inference time to quantify the level of trust in the predictions, in a single shot.
At its core our method uses error propagation \cite{taylor1997introduction}, commonly used in physics, where the error is equivalent to the variance. We treat the noise injected in a neural network as errors on the activation values. By training with noise injection, the network implicitly learns to minimize the accumulated errors on the training data distribution, since larger errors correspond to large loss signals. To give an intuition for this, we consider a simple computational graph (see Fig. \ref{fig:simpleerrorpropagation}). Let A and B be independent random variables and let $C = f_1(A,B)$ and $D = f_2(A,B)$ be, possibly non-linear, functions of A and B. Knowing the mean and the variance $\sigma^2_{A/B}$ of A and B, we want to compute the variance of C and D. We apply error propagation, where: 
\begin{equation}\label{simpleexample}
\sigma^2_{C/D} = \left(\frac{\partial f_{1/2}}{\partial A}\right)^2\sigma^2_A + \left(\frac{\partial f_{1/2}}{\partial B}\right)^2\sigma^2_B
\end{equation}

Note that the partial derivatives only approximate the outcome given non-linear functions $f_{1/2}$. Let us assume now that we have another function $E = f_3(C,D)$. We cannot apply Eq. \ref{simpleexample} directly to determine the variance of E because C and D, unlike A and B, are not statistically independent. Thus we have to consider the full covariance matrix of C and D which we can obtain again by applying error propagation. We start with the covariance matrix $\Sigma_{A,B}$ of A and B. This is obviously a diagonal matrix with entries $\sigma^2_A$ and $\sigma^2_B$ on its main diagonal. Then we can approximate the covariance matrix over C and D by computing:
\begin{equation}\label{covarianceexample}
\Sigma_{C,D} = J^T\Sigma_{A,B}J
\end{equation}

Here J is the Jacobian of the vector-valued function $\vec{f} = (f_1(A, B), f_2(A, B))^T$. The final variance $\sigma^2_E$ of E is obtained by applying Eq. \ref{covarianceexample} again but considering that $f_3$ is not a vector-valued function. Thus one can exchange the Jacobian with the gradient $J = \nabla_{C,D} f_3(C,D) = (\frac{\partial}{\partial C}, \frac{\partial}{\partial D})^T f_3(C,D)$. 

This minimal example already illustrates all the tools we need to approximate the variance at the output layer of a neural network given some noise layer, such as dropout or batch-norm \cite{teye2018bayesian} by applying error propagation. In the following, we explain the propagation of covariance for specific parts of a neural network: noise layers, affine layers, and non-linearities. Afterwards, we simplify the above equations for the common setup of a convolution and ReLU activation. This is necessary for high dimensional feature spaces due to the size of the covariance matrix (which scales quadratically with the number of activations).

We explicitly do not model the impact of correlations between activations on the activation means as we explore the applicability of performing variance propagation using error progation in neural networks and leave it to future work to incorporate it into the computational framework.

\subsection{Noise Layer}

Now we derive the covariance matrix of a noise layer's activation values in a neural network. The following consideration is independent of the implementation of the noise layer. We assume independent noise across the nodes of the noise layer which is not a necessity but common practice. In the following, superscripts correspond to layers and subscripts to nodes within a layer.

Consider a neural network with $l$ layers and $N$ noise layers at positions $i \in [0, l]$, where $l=0$ denotes the input. Let the input to a given noise layer be described by a random vector $\vec{X}$ with covariance matrix $\Sigma^{i-1}$. Furthermore, let the random vector representing the noise be $\vec{Z} \in \mathbb{R}^n$ with a diagonal covariance matrix $\Sigma_{\vec{Z}}$, where the entries of $\vec{Z}$ are independent and the entries of $\vec{X}$ are generally \textit{dependent}. There are two ways how the noise is commonly injected: addition and element-wise multiplication of $\vec{X}$ and $\vec{Z}$. 

When the noise injection is represented by adding the random vectors the resulting covariance at layer $i$ is simply given by
\begin{equation}\label{secondnoiseinject_addition}
\Sigma^i = \Sigma^{i-1} + \Sigma_{\vec{Z}}
\end{equation}

When the noise injection resembles an element-wise multiplication of the random vectors $\vec{X}$ and $\vec{Z}$ (e.g. dropout), the covariance matrix at layer $i$ is given by \cite{goodman1960exact}
\begin{multline}\label{secondnoiseinject_multiplication}
\Sigma^i = \Sigma_{\vec{Z}\circ\vec{X}, \vec{Z}\circ\vec{X}} =  \\
\Sigma_{\vec{Z}} \circ \Sigma^{i-1} +
 E[\vec{Z}]E[\vec{Z}]^T \circ \Sigma^{i-1} + E[\vec{X}]E[\vec{X}]^T \circ \Sigma_{\vec{Z}}
\end{multline}
where $\circ$ is the Hadamard product. We refer to the supplementary material for a detailed derivation of this formula. 

For the special case of the first noise layer in the network, we can either model the input noise from prior knowledge (\ie sensor noise) or simplify it by assuming zero noise. In the latter case, the resulting covariance matrix will be diagonal given independent noise, resulting in:
\begin{equation}\label{firstnoiseinject}
\Sigma^i = diag(\sigma_0^2,...,\sigma_n^2)
\end{equation}
where $n$ is the dimensionality of the activation vector and $\sigma_i^2$ is the variance of activation $i$. The variance introduced by the regular dropout, which follows a Bernoulli distribution, is given by $p(1-p)a_i^2$, with $p$ defining the dropout rate and $a_i$ the mean activation of node i.

\subsection{Affine Layers and Non-Linearities}

After obtaining the covariance matrix of a noise layer one needs to propagate it to the output layer. This usually means applying a series of affine layers and non-linearities. Here, we detail the case of a fully-connected and convolutional layer. It is straightforward to apply Eq. \ref{covarianceexample}, given that for the transformation of an affine layer the Jacobian J equals the weight matrix W. The covariance matrix is therefore,
\begin{equation}\label{covtransform}
\Sigma^{i} = W\Sigma^{i-1}W^T
\end{equation}
This is an exact transformation which does not depend on the underlying distribution. For the non-linearities in a neural network we approximate the transformation by a first-order Taylor expansion. The covariance transformation at a non-linearity is then given by:
\begin{equation}\label{eq:1}
\Sigma^{i} \approx J\Sigma^{i-1}J^T
\end{equation}
The particular Jacobians of the activation functions used in our experiments (ReLU, sigmoid and softmax) can be found in the supplementary material where we also provide an analysis of the error introduced by the first-order Taylor expansion of the softmax activation function. 

\subsection{Special Case: Convolutional Layers combined with ReLU Activations}

\begin{figure*}[h]
\begin{center}
\begin{tabular}{c@{\hskip 3pt}c@{\hskip 3pt}c@{\hskip 3pt}c@{\hskip 3pt}c@{\hskip 3pt}c@{\hskip 3pt}c@{\hskip 3pt}c@{\hskip 3pt}c}
    \multicolumn{9}{c}{\includegraphics[width=120mm]{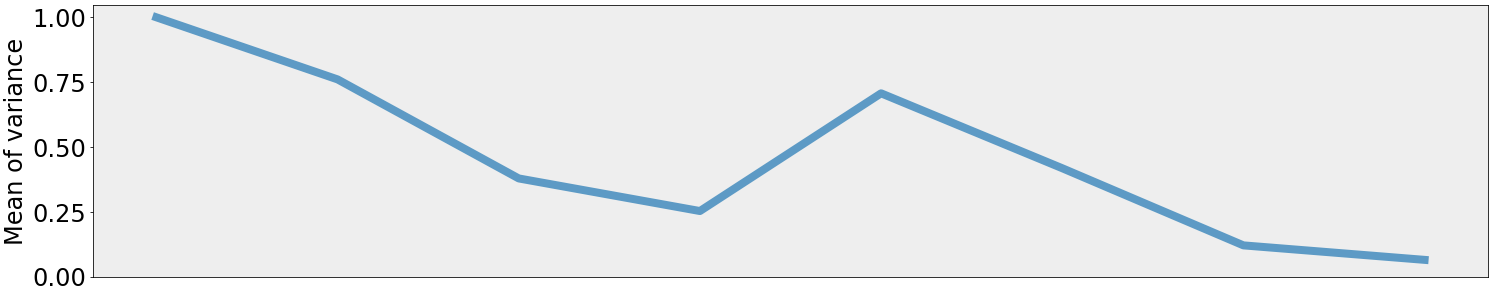}} \\
    Layer: &
    D& 
    C + R& 
    C + R& 
	C + R&     
	D& 
    C + R&
    C + R&
    C + R\\
    &
    \includegraphics[width=14mm]{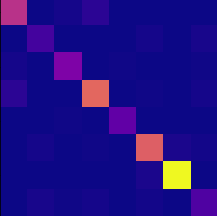}& 
    \includegraphics[width=14mm]{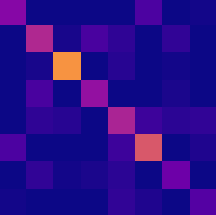}& 
    \includegraphics[width=14mm]{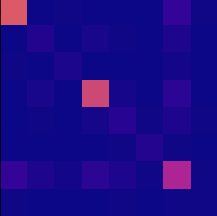}&
    \includegraphics[width=14mm]{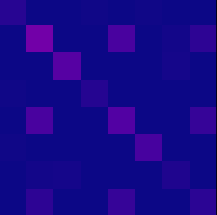}&
    \includegraphics[width=14mm]{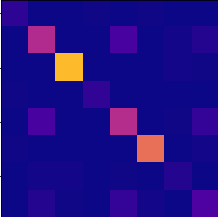}&
    \includegraphics[width=14mm]{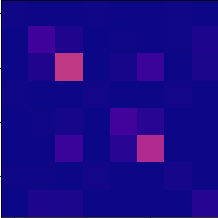}&
    \includegraphics[width=14mm]{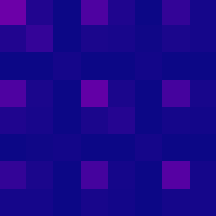}&
    \includegraphics[width=14mm]{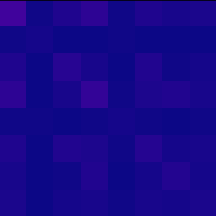}\\
\end{tabular}
\caption{We train a neural network (convolution - DropoutBlock - fully connected layer, where DropoutBlock corresponds to above sequence of dropout (D), convolution (C) and ReLU (R)) on CIFAR10. The feature maps in the DropoutBlock have the dimensionality 10x10x3 and the convolutional filters are of dimensionality 3x3x3. The upper part shows the mean variance of the activation values normalized with respect to the first dropout layer (blue). The lower part shows images (zoomed in for better visualization) of the mean covariance matrix at the corresponding layer (blue: small absolute values, yellow: large absolute value). We make two observations. Firstly, dropout strengthens the main diagonal. Secondly, variance decays in absence of additional dropout layers. Consequently, large regions of the covariance matrix stay approximately zero.} \label{example_var_decay}
\end{center}
\end{figure*}

Though the proposed approach does not require sampling, propagating the full covariance matrix may become prohibitively expensive for very high dimensional problems, such as images. This can be understood by considering that our method requires the full covariance matrix $\Sigma \in \mathbb{R}^{N \times N}$ at each layer with N nodes. This leads to a memory complexity of $O(N^2)$.  Since the many neural network architecture contains iterative applications of convolutional layers and rectified linear units, we simplify the above formulas for the sake of computational efficiency. 

For a fully connected layer, since each of its output nodes is a linear combination of all input nodes, modeling the full covariance is a necessity. However, this is not the case for a convolutional layer which strength comes from sharing weights across the entire input space and consequently applying a linear transformation only to a local neighborhood of each pixel. For the following approximation we assume a convolutional layer with kernel $K\in\mathbb{R}^{W^{'} \times H^{'} \times C^{'}}$ and an input to the convolutional layer  $I\in \mathbb{R}^{W \times H \times C}$ with $W^{'}\ll W$ and $H^{'}\ll H$. Given a diagonal covariance matrix prior to the convolutional layer, by applying e.g. a dropout layer, the output covariance would be a sparse matrix with a few non-zero entries off the main diagonal for local neighbourhoods in the input space. This is illustrated in Fig. \ref{example_var_decay} using a convolutional architecture on CIFAR10. Furthermore, given approximately symmetrical distributed weights ReLU activation leads to a probability of roughly 0.5 with which a variance value is dropped. This results in the observed decrease in the mean variance with the number of convolutional layers using ReLU activation functions. The observation that this architectural setup does not transport significant mass to wide regions of the covariance matrix motivates us to assume a \textit{diagonal covariance matrix}. 

As a result the computational complexity of propagating the variance reduces to the same level of normal forward propagation because one only needs to propagate the vector of the main diagonal of the covariance matrix. Under this assumption Eq. \ref{secondnoiseinject_multiplication} simplifies to:
\begin{multline}\label{secondnoiseinjectdiagcov}
V[\vec{X}^i] = E[\vec{X}^{i-1}]^2 \circ V[\vec{Z}] + \\ E[\vec{Z}]^2 \circ  V[\vec{X^{i-1}}] + V[\vec{X}^{i-1}] \circ  V[\vec{Z}]
\end{multline}

A derivation can be found in the supplementary material. To determine variances under the transformation of weight matrices and Jacobians of non-linearities using the above simplification, we need to square the corresponding matrix element-wise and multiply it with the vector representing the main diagonal of the covariance matrix: 
\begin{equation}\label{covtransform2}
Var[\vec{X}^i] = (W^2) Var[\vec{X}^{i-1}]
\end{equation}
and respectively
\begin{equation}\label{eq:2}
Var[\vec{X}^i] \approx (J^2) Var[\vec{X}^{i-1}]
\end{equation}

With the assumption of independent activations it is now straightforward to improve upon the simple Jacobian approximation of ReLU by explicitly computing the variance resulting from applying ReLU to a Gaussian. Therefore we assume a Gaussian distribution of activations prior to ReLU which is valid according to \cite{wang2013fast}. The respective formulas and derivations can be found in the supplementary material. 

\section{Experiments}

In this section we provide experimental evidence that our approximation can produce fast and accurate uncertainty estimates in a classification and regression setting.

\subsection{Synthetic Data}

First we want to give an intuition for the validity of our approach by applying it to a synthetic dataset, and comparing it directly to MC sampling. We create a regression dataset that consists of a single input and a single output, where the input is uniformly distributed in the interval $[0,20]$ and the target is the sine of the input plus Gaussian noise of mean $\mu=0$ and sigma $\sigma=0.3$. We fit a fully connected neural network with three hidden layers each containing 100 hidden units to the data. We apply dropout ($p=0.1$) prior to the last hidden layer. Since our primary goal is to approximate the epistemic uncertainty, we populate the test set with out-of-distribution samples (i.e. samples smaller than $0$ or larger than $20$) alongside samples from the training data distribution $[0,20]$. We approximate the variance, and thus the standard deviation, of the prediction in two ways - MC sampling (100 samples) and propagating the variance with our approximation.

\begin{figure}[h]
\centering
\begin{tabular}{c}
  \includegraphics[width=60mm]{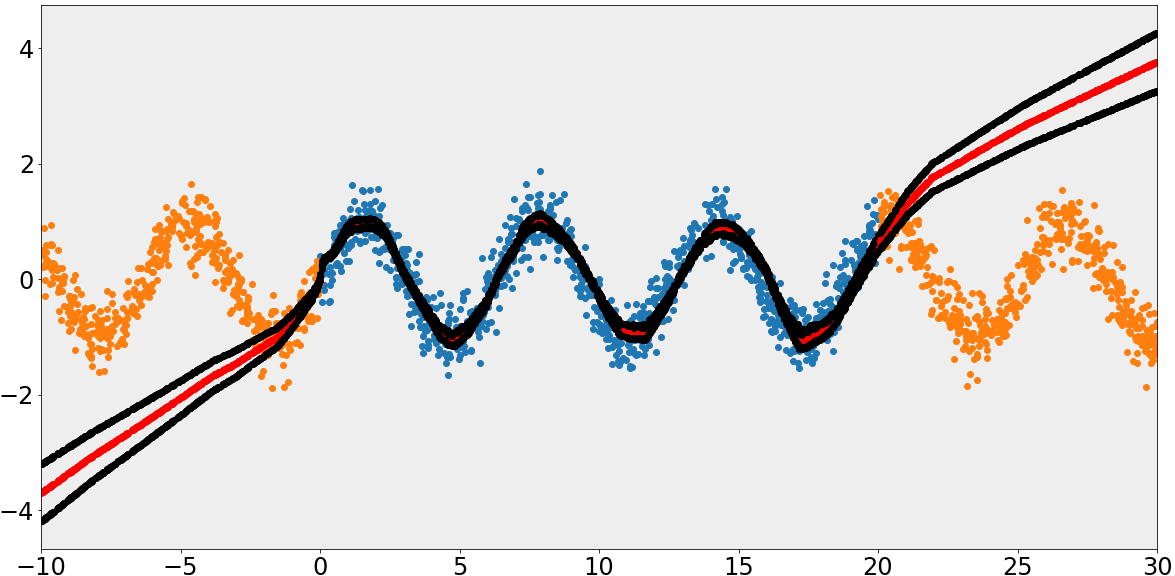}\\ 
  \includegraphics[width=60mm]{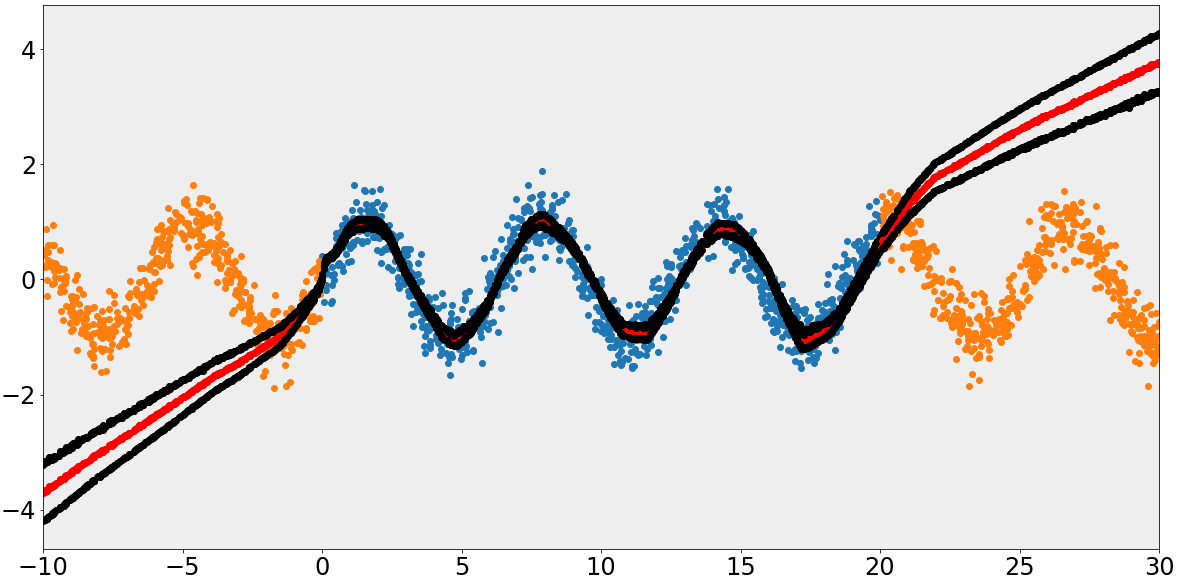}
\end{tabular}
\caption{Synthetic data. The neural network has three hidden layers with each 100 hidden units. We drop units prior to the last layer with probability $p=0.1$. We plot samples from the training data distribution (blue), samples outside of the training data distribution (orange), the prediction (red) and the prediction plus/minus the standard deviation (black). Upper: The standard deviation is approximated following our approach. Lower: The standard deviation is determined using MC dropout \cite{gal2016dropout} with 100 samples.}\label{toy_data_fit}
\end{figure}

Fig. \ref{toy_data_fit} visualizes the results for both approaches. Characteristically for epistemic uncertainty, we observe that the standard deviation of the prediction outside of the training data distribution increases. For this example, our approximation is in fact exact. In the supplementary material we show empirically that the sampling-based variance estimate converges towards our analytic estimate for large number of samples.

\subsection{Predictive Performance}

\begin{table*}[t]
\begin{center}
 \begin{tabular}{c | c c | c c | c c} 
 &\multicolumn{2}{c}{Test RMSE}&\multicolumn{2}{c}{Test log-likelihood}&\multicolumn{2}{c}{Runtime [s]} \\
 Dataset & MC\cite{gal2016dropout} & OUR &  MC\cite{gal2016dropout} & OUR &  MC\cite{gal2016dropout} & OUR \\
 \hline
 Boston Housing & $\mathbf{3.06\pm0.18}$ & $3.13 \pm 0.22$ & $\mathbf{-2.55 \pm 0.07}$ & $-2.65 \pm 0.12$ & 3.47 & \textbf{0.06}\\
 Concrete Strength & $\mathbf{5.42\pm0.10}$ & $\mathbf{5.42 	 \pm 0.11}$ & $\mathbf{-3.11 \pm 0.02}$ & $-3.13 \pm 0.02$ & 3.63 & \textbf{0.06} \\
 Energy Efficiency & $1.60\pm0.05$ & $\mathbf{1.59 \pm 0.05}$ & $\mathbf{-1.91 \pm 0.03}$ & $-1.96 \pm 0.03$ & 3.27 & \textbf{0.06}\\
 Kin8nm & $\mathbf{0.08 \pm 0.00}$ & $\mathbf{0.08 \pm 0.00}$ & $\phantom - 1.10 \pm 0.01$ & $\phantom - \mathbf{1.11 \pm 0.01}$ & 4.75 & \textbf{0.06}\\
 Naval Propulsion & $\mathbf{0.00\pm0.00}$ & $\mathbf{0.00 \pm 0.00}$ & $ \phantom - \mathbf{4.36 \pm 0.01}$ & $ \phantom - 3.64 \pm 0.02$ & 5.10 & \textbf{0.06}\\ 
 Power Plant & $\mathbf{4.04\pm0.04}$ & $4.05 \pm 0.04$ & $\mathbf{-2.82 \pm 0.01}$ & $-2.85 \pm 0.01$ & 4.46 & \textbf{0.06}\\
 Protein Structure & $\mathbf{4.42 \pm 0.03}$ & $\mathbf{4.42 \pm 0.03}$ & $\mathbf{-2.90 \pm 0.01}$ & $\mathbf{-2.90 \pm 0.00}$ & 4.38 & \textbf{0.06}\\
 Wine Quality Red & $\mathbf{0.63\pm0.01}$ & $\mathbf{0.63 \pm 0.01}$ & $\mathbf{-0.95 \pm 0.02}$ & $\mathbf{-0.95 \pm 0.01}$ & 3.49 & \textbf{0.06}\\
 Yacht Hydrodynamics & $\mathbf{2.89\pm0.25}$ & $3.14 \pm 0.31$ & $-2.32 \pm 0.10$ & $\mathbf{-2.10 \pm 0.07}$ & 3.42 & \textbf{0.06}\\
 \hline
\end{tabular}
\caption{table}{RMSE, test log-likelihood and runtime for MC dropout (MC) \cite{gal2016dropout} and our approximation (OUR). The error, denoted by $\pm$, is the standard error. For the RMSE smaller values are better, for the TLL larger values. We follow the original setup in \cite{gal2016dropout} and use T=10000 samples with MC dropout. Our runtime is 0.06 in every row due to decimal precision.}\label{uciresultstable}
\end{center}
\end{table*}

Following prior work estimating epistemic uncertainty \cite{hernandez2015probabilistic,gal2016dropout,teye2018bayesian} we analyze the predictive performance on 9 of the orignal 10 UCI regression datasets. We leave out the Year Prediction MSD dataset as done by other recent work \cite{teye2018bayesian}. We only compare our approach directly to MC dropout since we approximate the latter. 

We evaluate two metrics - root mean squared error (RMSE) on the test set and test log-likelihood (TLL). We expect our RMSE to be higher than MC dropout, as dropout sampling is a better approximation than scaling activations \cite{gal2016dropout}. The TLL represents the value of interest in our experiments as it quantifies the quality of the predicted distribution. It measures the probability mass on the target without making assumptions on the underlying distribution. 

We follow the original setup in \cite{hernandez2015probabilistic} \footnote{Instead of using Bayesian optimization \cite{snoek2012practical} for hyper-parameter optimization we use grid search}. We split the training data 20 times randomly into training and validation set, except for the dataset Protein Structure where we use five splits, and perform a separate grid search for the hyperparameters dropout rate and $\tau$. Following \cite{hernandez2015probabilistic,gal2016dropout} we use one hidden layer with 50 hidden units, except for Protein Structure where we use 100 hidden units. Dropout is applied directly to the input and after the hidden layer and we train the network for 400 epochs. We use the full covariance matrix to propagate uncertainty, defined in Eq \ref{secondnoiseinject_multiplication}, \ref{covtransform}, \ref{eq:1}.

The TLL in the original experiment requires sampling from distributions of outputs. Since our method naturally just returns the parameters of a unimodal distribution over the outputs, we assume a Gaussian distribution and sample from it to compute the TLL. According to \cite{wang2013fast} this is a reasonable assumption given our hidden layer dimension. We perform the same grid search as for MC dropout. For both, MC dropout and our proposed approximation, we sample 10000 predictions to compute the TLL.

Table \ref{uciresultstable} shows the results of this experiment. The TLL of our approximation is almost for every regression dataset only marginally lower or even larger than the TLL obtained by MC dropout. It is only for the Naval Propulsion dataset much worse than the original sampling-based method. However, given a RMSE of $0.00 \pm 0.00$  we perfectly fit this dataset. In this case, extremely confident and accurate predictions may lead to a regime where the Gaussian assumption loses validity or higher accuracy of MC sampling has a stronger impact on the TLL. Given the deviation of other methods from MC dropout, which can be found in \cite{gal2016dropout}, our approximation performs well.

\subsection{Classification Task: Bayesian SegNet}

A recent large scale architecture applying MC dropout is Bayesian SegNet \cite{Kendall2015BayesianSM} for semantic segmentation. The original work examines several architectures differing in the placement of dropout within the network. According to the authors, for the quality of the uncertainty estimate the location of dropout within the architecture is irrelevant. We train consider Bayesian SegNet on CamVid dataset \cite{brostow2009semantic} and investigate two of their probabilistic variants:
\begin{itemize}
  \item Dropout after the central four encoder and decoder blocks (ENCDEC). This is the best performing setup in terms of prediction \cite{Kendall2015BayesianSM}. It is challenging for our approximation since the last dropout layer is placed far away from the output layer. Thus our methods naturally underestimates the variance of the prediction since we assume a diagonal covariance matrix.
  \item One dropout layer before the final classifier layer (CLASS). This setup fits better our method, since the dropout layer is closer to the output, this reducing the number of error sources. 
\end{itemize} 

We compare the performance of our implementation to the original work in Table \ref{ourbayesiansegnetperformance}. We emphasize that the goal of this work is not to outperform the state of art, but rather to provide an efficient sampling-free approximation of epistemic uncertainty. We thus aim at showcasing the similarity between sampling-based uncertainty estimation and the proposed approximation. 

\begin{table}[t]
\begin{center}
 \begin{tabular}{c | c | c | c} 
  Method & G & C & I/U \\
  \hline
  Original Bayesian SegNet\cite{Kendall2015BayesianSM} & \textbf{86.9} & 76.3 & \textbf{63.1} \\ [0.5ex] 
  Our implementation & 86.1 & \textbf{76.4} & 54.1 \\ [0.5ex] 
 \hline 
\end{tabular}
\caption{table}{We compare our retrained Bayesian SegNet (ENCDEC architecture) \cite{Kendall2015BayesianSM} to the results presented in the original work. We compare the quantities global accuracy (G), average class accuracy (C) and mean intersection over union.}\label{ourbayesiansegnetperformance}
\end{center}
\end{table}

We trained with batch size 8 and stochastic gradient descent with initial learning rate $0.1$ and exponential learning rate decay with base $0.95$ for 200 epochs. We use early stopping (watching the validation loss) with patience 50. 
The original images in CamVid have resolution 720x960 and 32 classes. Following \cite{Kendall2015BayesianSM} we only use 11 generalized classes and downsample the images to 360x480.
 
Since semantic segmentation is a classification task, one obtains variances for each class at each pixel. There exist several strategies to aggregate these variance into a scalar quantity (for a selective list see \cite{gal2017deep}). Subsequently we follow the original work \cite{Kendall2015BayesianSM} and use the mean standard deviation of the softmax scores at each pixel. We apply our approximation according to Eq. \ref{secondnoiseinjectdiagcov}, \ref{covtransform2}, \ref{eq:2}. 

\begin{figure}[h]
\centering
\begin{tabular}{c@{\hskip 3pt}c@{\hskip 3pt}c}
  \includegraphics[width=20mm]{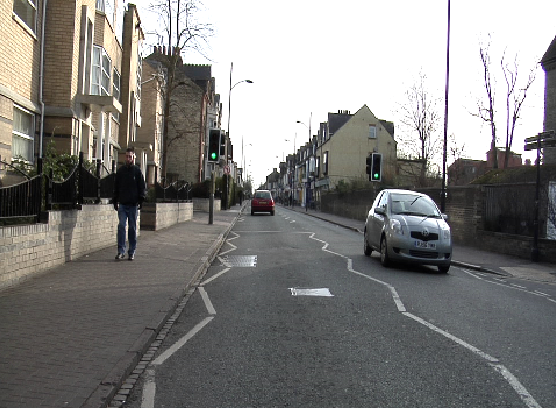} &   \includegraphics[width=20mm]{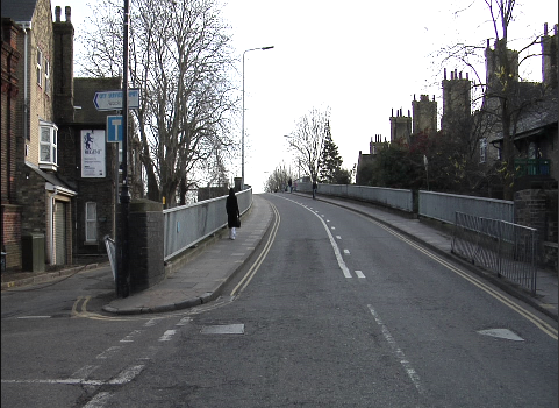} &  \includegraphics[width=20mm]{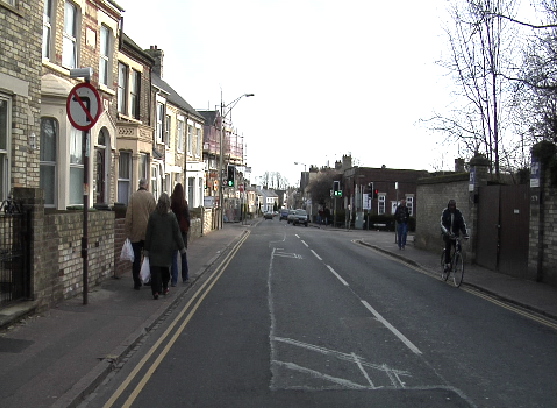}\\ 
  \includegraphics[width=20mm]{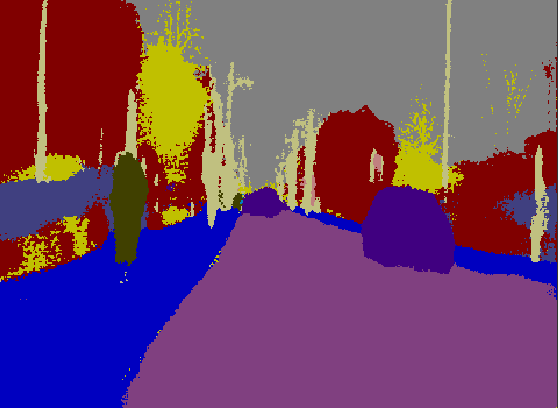} & 
  \includegraphics[width=20mm]{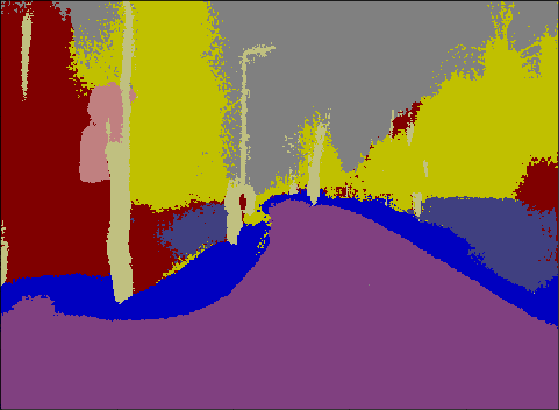} & 
  \includegraphics[width=20mm]{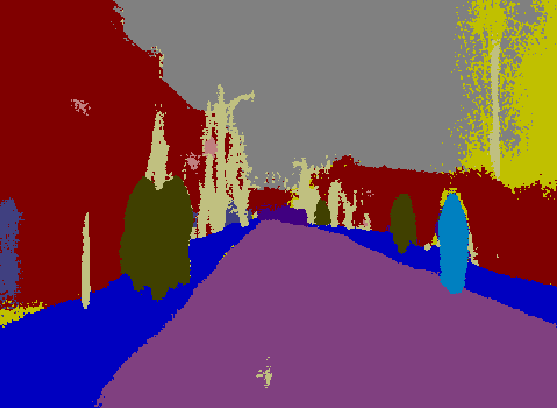}\\
  \includegraphics[width=20mm]{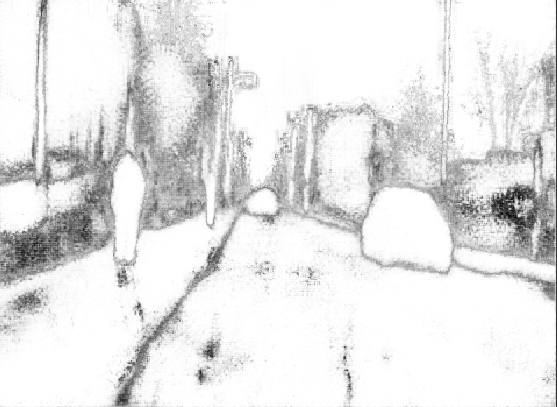} & \includegraphics[width=20mm]{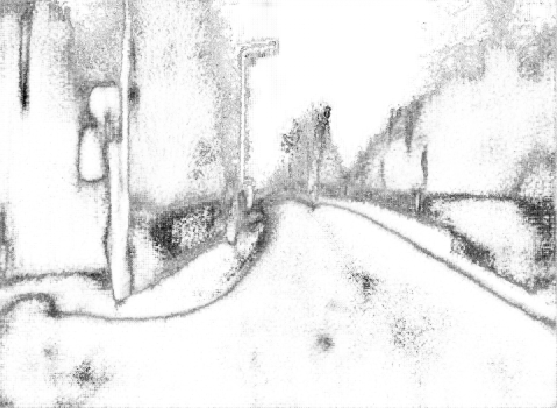} & \includegraphics[width=20mm]{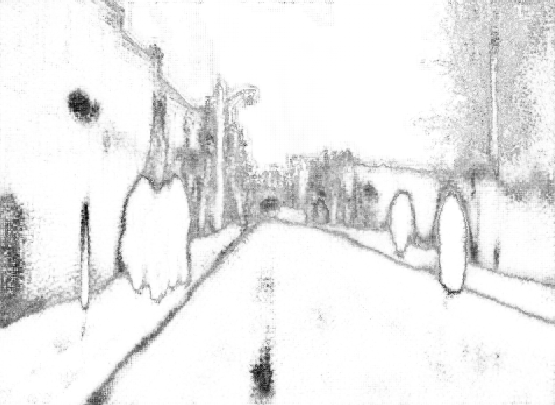}\\
  \includegraphics[width=20mm]{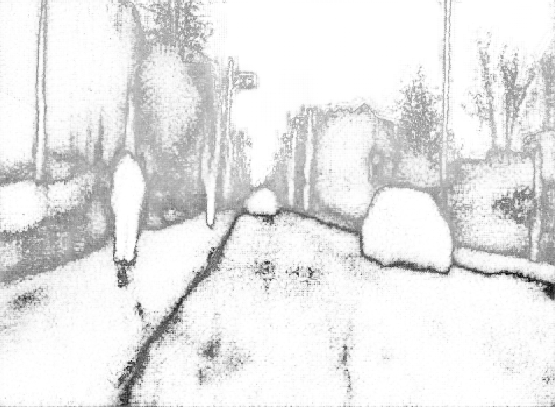} & \includegraphics[width=20mm]{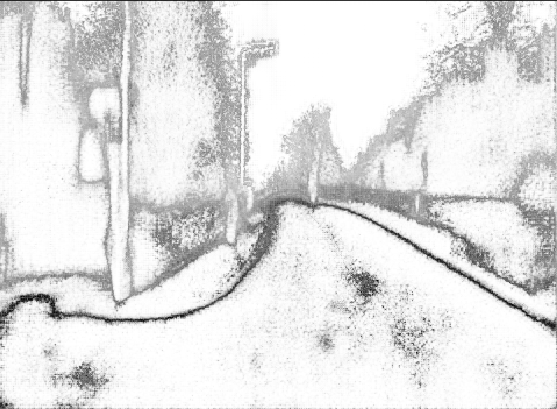} & \includegraphics[width=20mm]{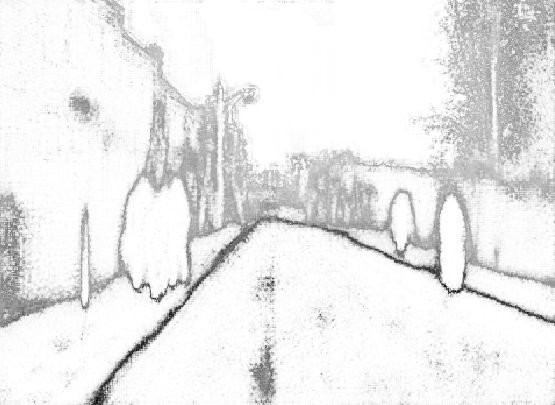}\\
\end{tabular}
\caption{Qualitative results of our approximation. First row: Input image. Second row: Segmentation result using MC dropout \cite{gal2016dropout} with 50 samples. Third row: Uncertainty estimate using MC dropout. Fourth row: Our approximation. We use differently scaled colormaps for the uncertainty maps to emphasize their structural similarity.}\label{qualitative_bayesian_encdec}
\end{figure}

\begin{figure}[h]
\begin{tabular}{c@{\hskip 3pt}c}
	a) Misclassification rate & b) Runtime comparision \\
  \includegraphics[width=40mm]{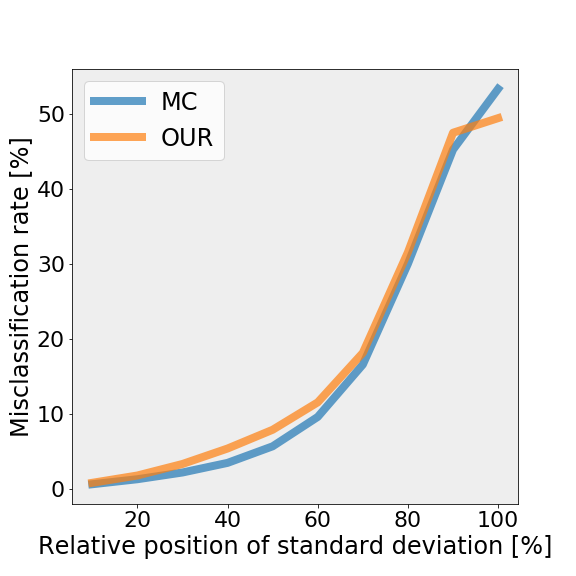} & 
  \includegraphics[width=40mm]{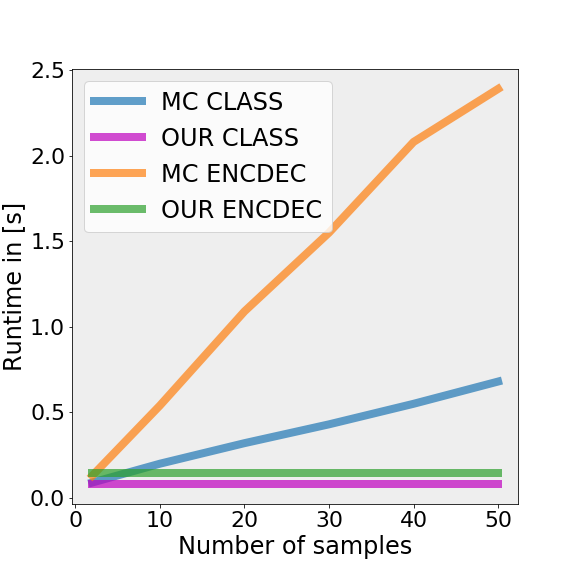}\\
\end{tabular}
\caption{a): Comparison of pixel misclassification rate depending on quantile of standard deviation between MC dropout \cite{gal2016dropout} with 50 samples (blue, MC) and our approximation (orange, OUR) (e.g. $50\%$ implies that $50\%$ of the pixels have a smaller standard deviation.)
b): Runtime comparison between MC dropout and our approximation (OUR) for both architectures (ENCDEC and CLASS). Our approximation (ENCDEC: green, CLASS: red) is constant (no sampling) and MC dropout (ENCDEC: orange, CLASS: blue) increases linearly with the number of samples. We cache results prior to the first dropout layer to optimize performance of MC dropout.}\label{runtime_misclassification_rate_SegNet}
\end{figure}

Fig. \ref{qualitative_bayesian_encdec} and \ref{teaser_classArchitecure_uncertainty} show qualitative results of our approximation and the sampling-based estimates (a video can be found in the supplementary material). We first approximate the uncertainty estimates using the ENCDEC architecture. For all qualitative results we only show the image, MC prediction, MC uncertainty estimate and our approximation, since our work focuses on producing sampling-free uncertainty estimates. For qualitative examples including our prediction and the groud truth we refer to the supplementary material. Both methods predict similar regions of high uncertainty. Those are mainly the object boundaries resulting from the noise induced by human labelers. Given the qualitative similarity of the predicted uncertainty, we surprisingly found that the magnitude of the approximated uncertainty is much lower than the sampling-based uncertainty (mean absolute difference is 93.7\% of the mean variance received by MC sampling). We understand that the difference arises from the fact that in this architecture the last dropout layer is far away from the output layer. Given the qualitative similarity of the predicted uncertainty, we deduct that mixture terms, which lie off the main diagonal of the covariance matrix, primarily act as a variance bias. 

Fig. \ref{runtime_misclassification_rate_SegNet} compares the runtime of sampling-based uncertainty estimation with our approximation. We cache the results prior to the first dropout layer in the architecture and only repeatedly propagate the part of the network that contains dropout layers to optimize MC sampling. As expected we observe a linear dependence on the number of samples of the sampling-based approach where the slope depends on the location of the first dropout layer. We clearly see the computational advantage of our approximation.

To prove coherence of the predicted uncertainties on the training data distribution, we investigate the correlation of the uncertainty with the misclassification rate on the test set. Since we are explicitly interested in epistemic uncertainty, we further prove that our uncertainty estimate increases for out-of-distribution samples. 

We plot the pixel misclassification rate against the uncertainty value (see Fig. \ref{runtime_misclassification_rate_SegNet}). Obviously we wish to observe an increasing pixel misclassification rate with larger uncertainty estimates. Since the scale of magnitude of our uncertainty estimates is different from the sampling-based approach we do not plot it directly against the uncertainty value. To be able to compare their behavior directly, we plot the misclassification rate against quantile of the uncertainty estimate. Both curves behave similarly which implies similar calibration of the uncertainty.

Moreover, we use the CLASS architecture to investigate whether our approximation is able to detect out-of-distribution samples. It is difficult to design adequate experiments to validate this capability of an epistemic uncertainty estimate. Here, we withhold certain classes during training time and present them to the network at test time. This way we exclude respective regions of the training data distribution. We exclude two classes: pedestrians and cyclists. We selected them based on the fact that they differ from other classes in their appearance, are similar to each other and do not make up large enough areas of the images to endanger convergence of the training. Qualitative results of this training can be found in the supplementary material. Fig \ref{uncertainty_per_class_hold_out} shows that the mean uncertainty value of each classes with and without withholding classes. The average relative increase of uncertainty is maximal for withheld classes.

\begin{figure}[h]
\centering
  \includegraphics[width=80mm]{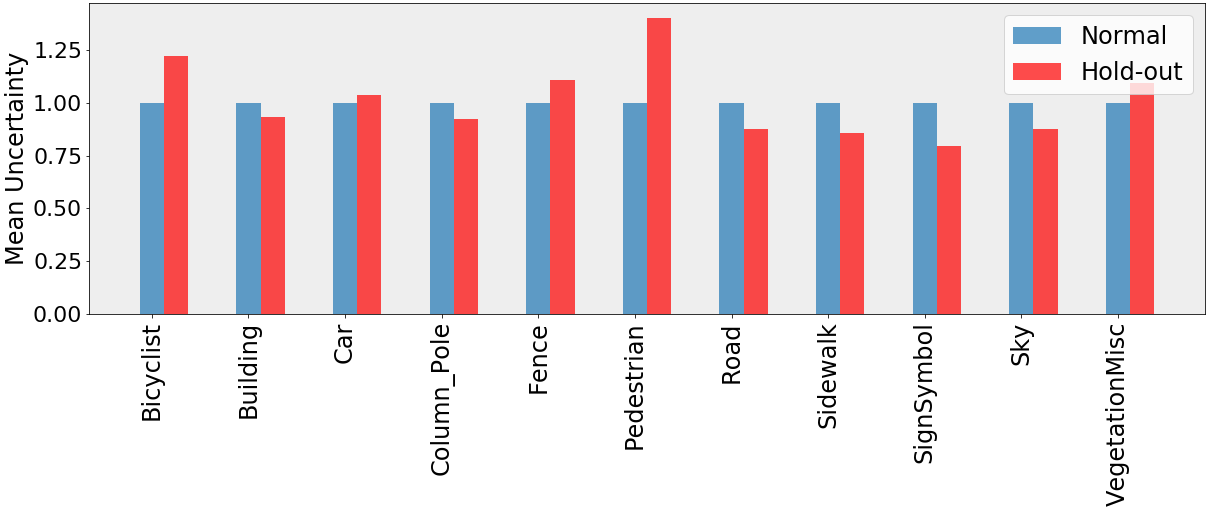}
\caption{Mean uncertainty (using our approximation) per class for CLASS architecture using all classes (blue) and withholding the classes for pedestrians and cyclists (red). Uncertainties are normalized to the uncertainty value observed using all classes to highlight their relative change. The absolute value strongly depends on the boundary to area ratio of each class since high uncertainties mostly occur at boundaries.}\label{uncertainty_per_class_hold_out}
\end{figure}

\subsection{Regression Task: Depth Regression}

We now evaluate our approach on a regression task. Without a softmax, we expect our approximation to not only show similar behavior but also to be compellingly close to the sampling-based approach. 
We apply our approximation to monocular depth regression \cite{godard2017unsupervised}. The final activation of this architecture is a sigmoid function. Thus it is expected that our approximation will not exaclty match the sampling-based result. Following the original work we train on the KITTI dataset \cite{geiger2013vision} and keep the setup unchanged with exception of inserting a dropout layer prior to the final convolution to estimate uncertainty. As uncertainty estimate we choose the variance of the regression output and we use Eq. \ref{secondnoiseinjectdiagcov}, \ref{covtransform2} and \ref{eq:2} to propagate variance. Note that this setup does not require modeling the full covariance matrix because sigmoid is an element-wise operation. 

\begin{figure}[h]
\centering
\begin{tabular}{c@{\hskip 3pt}c@{\hskip 3pt}c}
  \includegraphics[width=20mm]{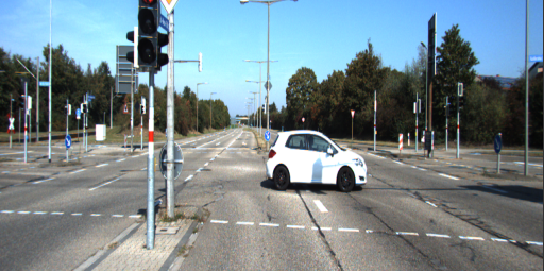} &   
  \includegraphics[width=20mm]{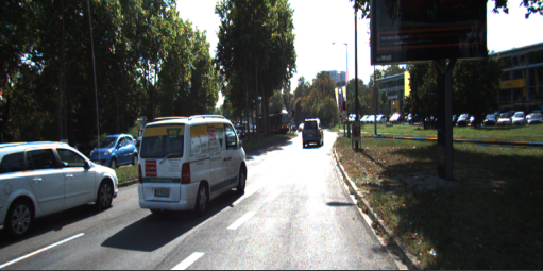} &  	  
  \includegraphics[width=20mm]{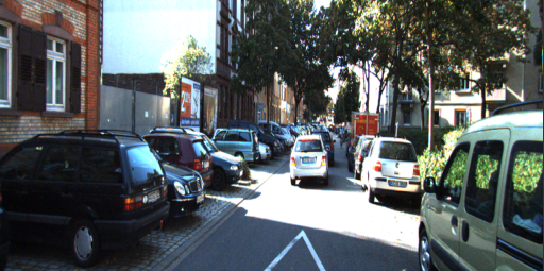}\\ 
  \includegraphics[width=20mm]{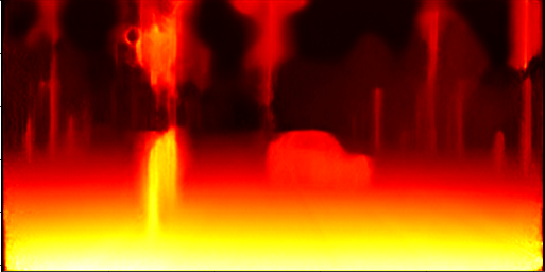} &   
  \includegraphics[width=20mm]{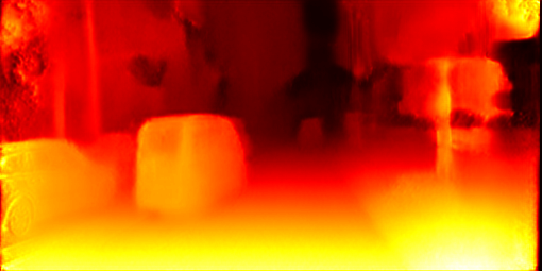} &  	  
  \includegraphics[width=20mm]{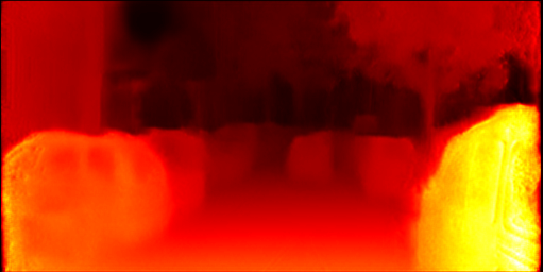}\\  
  \includegraphics[width=20mm]{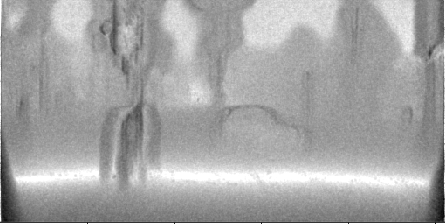} & 
  \includegraphics[width=20mm]{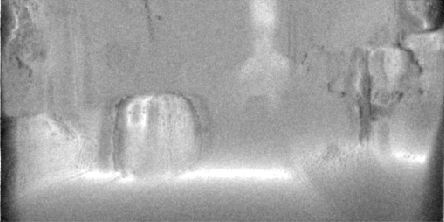} &  	  
  \includegraphics[width=20mm]{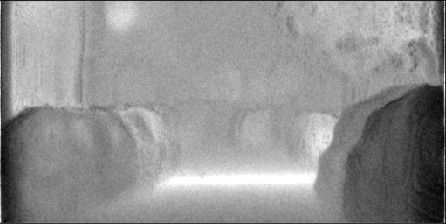}\\ 
  \includegraphics[width=20mm]{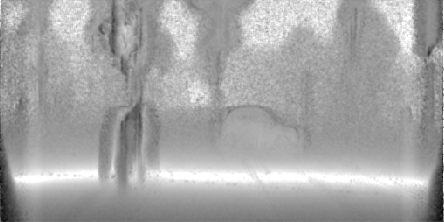} &   
  \includegraphics[width=20mm]{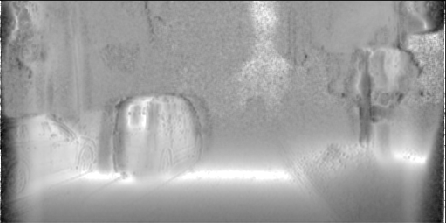} &  	
  \includegraphics[width=20mm]{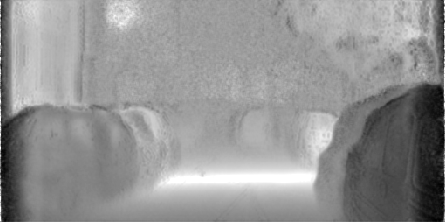}\\ 
\end{tabular}
\caption{Qualitative results of our depth regression network. First row: Input image. Second row: Depth regression using MC dropout \cite{gal2016dropout}. Third row: Logarithm of variance (white: low, black: high) using MC dropout with 50 samples. We use the logarithm because high uncertainties from non-overlapping stereo images on the left and right of the image are dominating the colormap. Fourth row: logarithm of variance using our approximation.}\label{qualitative_monodepth}
\end{figure}

\begin{figure}[h]
\begin{tabular}{c@{\hskip 3pt}c}
	a) Misclassification rate & b) Runtime comparision \\
  \includegraphics[width=37mm]{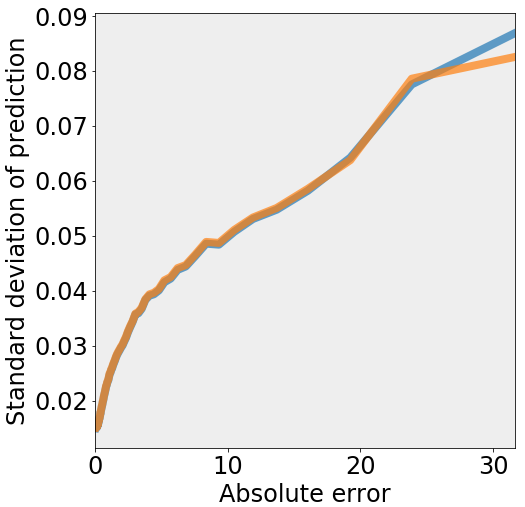} &
  \includegraphics[width=43mm]{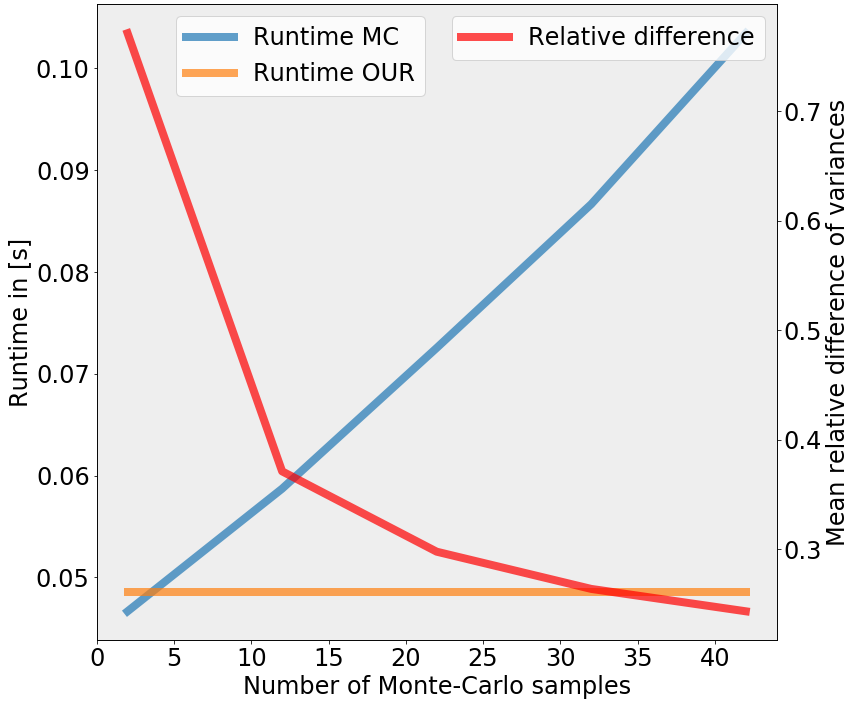} \\
\end{tabular}
\caption{a): Standard deviation of depth prediction depending on the absolute difference between the predicted depth and the ground truth. The standard deviation for both, MC dropout \cite{gal2016dropout} with 50 samples (blue) and our approximation (orange), behave similarly.
b): Runtime of our method (orange) and MC dropout \cite{gal2016dropout} with 50 samples (blue) depending on the number of samples. We visualize the mean relative difference between our variance approximation and the variance computation based on MC dropout (red).}\label{runtime_misclassification_rate_monodepth}
\end{figure}

Fig. \ref{qualitative_monodepth} shows qualitative results of this experiment. We visualize the logarithm of the uncertainty value since high uncertainty at the left and the right border is dominating the colormap. This is an artifact of the uncerlying training method. Given stereo pairs of images the network learns to predict disparity maps transforming one image into the other. Due to non-overlapping borders, reconstruction fails in those regions and consequently the uncertainty is high. Furthermore we consistently observe an uncertainty minimum at a certain depth. This is the area of optimal depth resolution of the stereo camera setup. The depth resolution of a point in space increases quadratically with its distance from the camera. On the contrary the depth resolution is proportional to the disparity error which increases for closer objects due to larger disparities.

We analyze how the mean absolute variance difference relative to the mean variance and runtime depend on the number of MC samples. The result is displayed in Fig. \ref{runtime_misclassification_rate_monodepth} b). As expected the runtime of the sampling-based approach increases linearly with the number of samples. On the other hand the mean absolute difference decreases with the number of samples. This suggests that our approach is not only superior in terms of runtime but also in terms of the uncertainty estimate. The fact that the relative difference does not coverge to zero originates from the usage of the sigmoid activation function after the final layer.

Finally we show the meaningfulness of our approximated uncertainty predictions and the sampling-based results by plotting their correlation with the absolute difference between the predicted depth and the ground truth (see Fig. \ref{runtime_misclassification_rate_monodepth} a)). The standard deviation of both , the sampling-based result and our approximation of it, increase linearly with the absolute error. Thus our uncertainty estimate can be used to identify regions which are likely to have large regression errors while adding minimal computational overhead.

\section{Conclusion}

We have shown that the framework of error propagation can be used to approximate the sampling procedure of epistemic uncertainty estimates that rely on noise injection at training time. We applied the proposed approximation to two large scale computer vision tasks illustrating the computational efficiency and the coherence of the resulting uncertainty maps. The approximation is numerically better for noise layer located closer to the output layer. Having a methodology to analytically approximate the uncertainty estimate based on stochastic regularization, in future research we aim to represent the noise injection by a loss function which will enable us to learn the noise parameter (e.g. dropout rate). This has the potential to provide an estimate for \textit{aleatoric and epistemic} uncertainty.

\section{Acknowledgements}
We thank Nutan Chen and Seong Tae Kim for the valuable discussions and constructive feedback. We further thank Autonomous Intelligent Driving GmbH for supporting this work. 

{\small
\bibliographystyle{ieee}
\bibliography{egbib}
}

\title{Supplementary Material for "Sampling-free Epistemic Uncertainty Estimation Using Approximated Variance Propagation"}

\author{}

\date{}

\maketitle


Subsequently we provide additional information. In section \ref{section1} we write down the Jacobians which are used to approximately propagate the covariance matrix through non-linearities. Section \ref{section2} and \ref{section3} we derive the formula for the covariance matrix of the element-wise product of independent random vectors and, respectively, the expectation and variance of ReLU given a Gaussian distribution. Section \ref{comparable_runtime_experiment} compares the performance of Monte-Carlo dropout with similar runtime (few samples) with our work. Further, we show empirically in section \ref{section4} for that the sampling-based approach converges to our analytic form for the case of the synthetic dataset. Finally in section \ref{section5} we show further qualitative results of our experiments. We also refer to the video which can be found in the supplementary material for qualitative results. 

\section{Jacobians of Activation Functions}\label{section1}

We show the Jacobians of the activation functions - ReLU, sigmoid and softmax - that are used throughout our experiments. For ReLU we assume its derivative in the origin to be zero. Then the Jacobian is given by

\begin{equation}\label{relujacobian}
    J_{ij}(ReLU(\vec{x}))= 
\begin{cases}
    1,& \text{if } i=j\text{ and }x_{i} > 0\\
    0,              & \text{otherwise}
\end{cases}
\end{equation}

In case of an element-wise sigmoid function $\sigma(\vec{x})$ the Jacobian is given by

\begin{equation}\label{sigmoidjacobian}
    J_{ij}(\sigma(\vec{x}))= 
\begin{cases}
    \sigma(x_i)(1-\sigma(x_i)),& \text{if } i=j\\
    0,              & \text{otherwise}
\end{cases}
\end{equation}

and for the softmax respectively

\begin{equation}\label{softmaxjacobian}
J_{ij} = S_i(\delta_{ij} - S_j)
\end{equation}

where $S_i$ and $S_j$ are the i-th and j-th entry of the softmax output and $\delta_{ij}$ is the Kronecker delta.

\section{Covariance of Hadamard Product of Random Vectors}\label{section2}

In the following $X$ and $Z$ denote random variables and $\vec{X}$ and $\vec{Z}$ denote random vectors, which may each have a non-diagonal covariance matrix but do not depend on each other.  Further $\Sigma_{\vec{X}}$ denote the covariance matrix of $\vec{X}$. 

We are interested in the covariance matrix of $\vec{Y} = \vec{Z} \circ \vec{X}$ resulting from an element-wise multiplication of $\vec{Z}$ and $\vec{X}$. Therefore we plug $\vec{Z} \circ \vec{X}$ into the definition of the covariance matrix:

\begin{equation}\label{cov01}
\Sigma_{\vec{Z} \circ \vec{X}} = E[(\vec{Z} \circ \vec{X})(\vec{Z} \circ \vec{X})^T] - E[\vec{Z} \circ \vec{X}]E[\vec{Z} \circ \vec{X}]^T
\end{equation}

Given that $\vec{Z}$ and $\vec{X}$ are independent and that

\begin{equation}
(\vec{Z} \circ \vec{X})(\vec{Z} \circ \vec{X})^T = (\vec{Z}\vec{Z}^T) \circ (\vec{X}\vec{X}^T)
\end{equation}

Eq. \ref{cov01} yields:

\begin{multline}\label{cov03}
\Sigma_{\vec{Z} \circ \vec{X}} = E[\vec{Z}\vec{Z}^T] \circ E[\vec{X} \vec{X}^T] - 
\\
(E[\vec{Z}]E[\vec{Z}]^T) \circ (E[\vec{X}]E[\vec{X}]^T)
\end{multline}

Now we can compare Eq. \ref{cov03} with 

\begin{multline}\label{cov04}
\Sigma_Z \circ \Sigma_X = 
\\ 
(E[\vec{Z}\vec{Z}^T] - E[\vec{Z}]E[\vec{Z}]^T) \circ (E[\vec{X}\vec{X}^T] - E[\vec{X}]E[\vec{X}]^T) 
\\
= E[\vec{Z}\vec{Z}^T] \circ E[\vec{X}\vec{X}^T] + E[\vec{Z}]E[\vec{Z}]^T \circ E[\vec{X}]E[\vec{X}]^T - 
\\
E[\vec{Z}\vec{Z}^T] \circ E[\vec{X}]E[\vec{X}]^T -  E[\vec{X}\vec{X}^T] \circ E[\vec{Z}]E[\vec{Z}]^T
\end{multline}

and see that Eq. \ref{cov03} is equivalent to

\begin{multline}\label{cov06}
\Sigma_{\vec{Z} \circ \vec{X}} = \Sigma_Z \circ \Sigma_X +
\\
E[\vec{Z}\vec{Z}^T] \circ E[\vec{X}]E[\vec{X}]^T + E[\vec{X}\vec{X}^T] \circ E[\vec{Z}]E[\vec{Z}]^T -
\\
2(E[\vec{Z}]E[\vec{Z}]^T) \circ (E[\vec{X}]E[\vec{X}]^T) 
\\
= \Sigma_Z \circ \Sigma_X + E[\vec{X}]E[\vec{X}]^T \circ (E[\vec{Z}\vec{Z}^T] - E[\vec{Z}]E[\vec{Z}]^T) + 
\\
E[\vec{Z}]E[\vec{Z}]^T \circ (E[\vec{X}\vec{X}^T] - E[\vec{X}]E[\vec{X}]^T)
\end{multline}

Using the definition of the covariance matrix we obtain the desired form of the equation:

\begin{multline}\label{cov07}
\Sigma_{\vec{Z} \circ \vec{X}} = \Sigma_Z \circ \Sigma_X +  E[\vec{X}]E[\vec{X}]^T \circ \Sigma_Z + 
\\
E[\vec{Z}]E[\vec{Z}]^T \circ \Sigma_X
\end{multline}

\section{Expectation and Variance of ReLU Given a Gaussian Distribution}\label{section3}

We write down the first- and second-order moments of a $f(X) = max(0, X)$ where X is a scalar, normal distributed random variable. We are only interested in scalar inputs since we write out these formulars for the assumption of a diagonal covariance matrix.

Given a univariate Gaussian $N(\mu, \sigma)$ with mean $\mu$ and standard deviation $\sigma$, the expectation of $f(X)$ is determined by the integral

\begin{multline}\label{ReLU_exp}
E_{X \propto N(\mu, \sigma)}[max(0, X)] \\ 
= \frac{1}{\sqrt{2\pi}\sigma}\int_{0}^{\infty} x exp\left( -\frac{(x-\mu)^2}{2\sigma^2} \right) dx \\
= \sqrt{\frac{1}{2\pi}} \sigma exp\left( -\frac{\mu^2}{2\sigma^2} \right) + \frac{\mu}{2} \left( 1 - erf\left( \frac{\mu}{\sqrt{2}\sigma} \right) \right)
\end{multline}

where $erf(x)$ is the error function with

\begin{equation}\label{errf}
erf(x) = \frac{2}{\sqrt{\pi}} \int_{0}^{x} exp(-z^2)dz
\end{equation}

The variance is then given by

\begin{multline}\label{ReLU_exp}
Var_{X \propto N(\mu, \sigma)}[max(0, X)] \\ 
= E_{X \propto N(\mu, \sigma)}[max(0, X)^2] - E_{X \propto N(\mu, \sigma)}[max(0, X)]^2
\end{multline}

Here we know $E_{X \propto N(\mu, \sigma)}[max(0, X)]^2$ via the expectation. The other term yields 

\begin{multline}\label{ReLU_exp}
E_{X \propto N(\mu, \sigma)}[max(0, X)^2] \\ 
= \frac{1}{\sqrt{2\pi}\sigma}\int_{0}^{\infty} x^2 exp\left( -\frac{(x-\mu)^2}{2\sigma^2} \right) dx \\
= \frac{1}{2}\left( \sigma + \mu^2 \right) \left( 1 + erf\left( \frac{\mu}{\sqrt{2}\sigma} \right) \right) + \frac{\mu \sigma}{\sqrt{2\pi}} exp\left( -\frac{\mu^2}{2\sigma^2} \right)
\end{multline}

\section{Comparison with Monte-Carlo (MC) Dropout of Similar Computational Cost}\label{comparable_runtime_experiment}

\begin{figure}[t]
\begin{center}
\begin{tabular}{c@{\hskip 3pt}c}
	 \includegraphics[width=4cm]{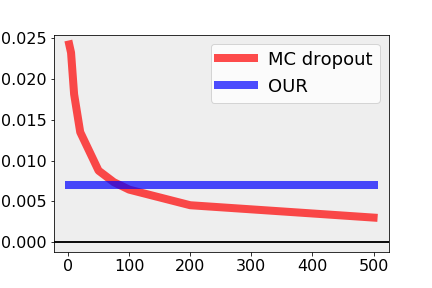} &
	 \includegraphics[width=4cm]{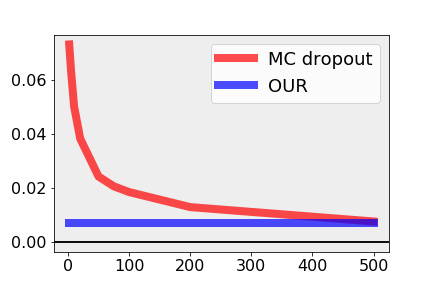}\\
\end{tabular}
\end{center}
   \caption{Mean absolute difference of MC dropout/OUR and GT (red/blue) depending on the number of samples on Boston Housing. OUR is constant without sampling. Left: Mean. Right: STD.}
   \vspace{-10pt}
\label{fig:comparable_runtime}
\end{figure}

We evaluate the predicted mean and standard deviation (STD) of our approach (OUR) and MC dropout on Boston Housing treating the result with 10000 samples as ground truth (GT). OUR is of computational advantage for more than approximately 175 samples. Fig. \ref{fig:comparable_runtime} shows the Mean absolute difference of MC dropout/OUR and GT (red/blue) depending on the number of samples. Even for up to 500 samples MC dropout fails to match the accuracy of our STD approximation. The mean approximation of MC dropout performs already better for much fewer samples ($>$100). 

\section{Absolute Variance Difference for Synthetic Data}\label{section4}

\begin{figure}[h]
    \centering
    \includegraphics[width=0.30\textwidth]{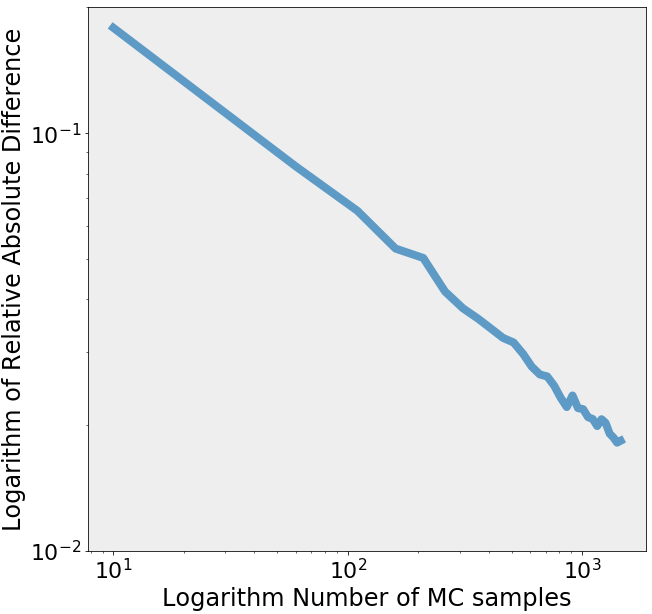}
    \caption{Relative absolute error between standard deviation obtained by Monte-Carlo dropout \cite{gal2016dropout} and our approximation in a double logarithmic plot. We observe that the relative absolute difference approaches increasingly small values for larger numbers of samples.}
    \label{fig:RelativeErrorToyData}
\end{figure}

We fit a neural network to a synthetic dataset. In \ref{fig:RelativeErrorToyData} we show empirically that the sampling-based variance estimate converges to our analytic expression. We observe that the relative absolute difference between the sampling-based variance estimate and our approximation converges to zero for large numbers of samples.

\section{Qualitative Results Including Our Prediction}\label{section5}

\subsection{Bayesian SegNet \cite{Kendall2015BayesianSM}}

\begin{figure*}[h]
\begin{center}
\begin{tabular}{c@{\hskip 3pt}c@{\hskip 3pt}c@{\hskip 3pt}c@{\hskip 3pt}c}
\includegraphics[width=26mm]{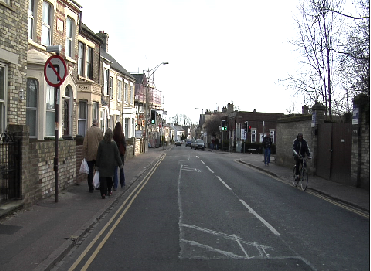}&   \includegraphics[width=26mm]{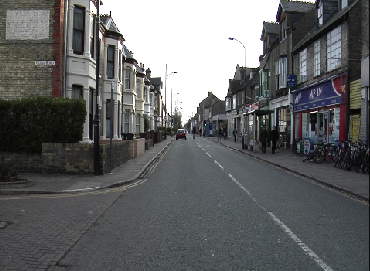}&   \includegraphics[width=26mm]{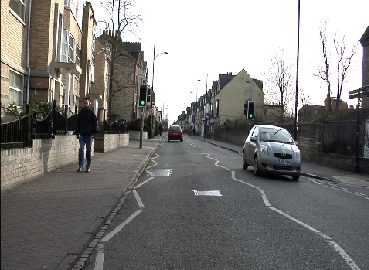}&  \includegraphics[width=26mm]{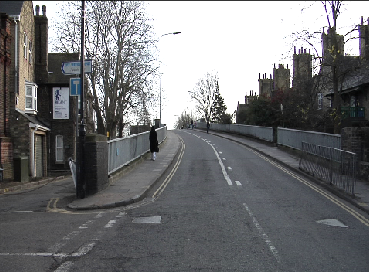}&  \includegraphics[width=26mm]{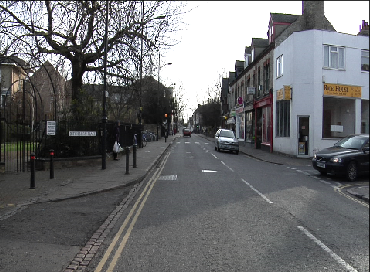}\\ 
\includegraphics[width=26mm]{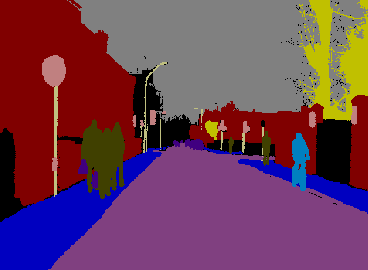} &   \includegraphics[width=26mm]{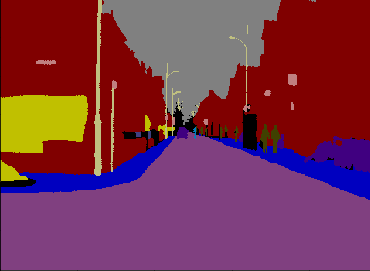} &   \includegraphics[width=26mm]{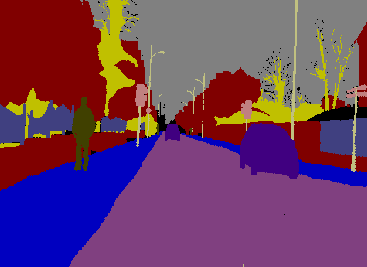} &  
\includegraphics[width=26mm]{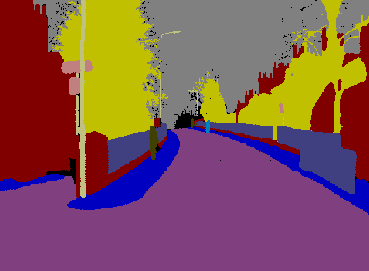} &  
\includegraphics[width=26mm]{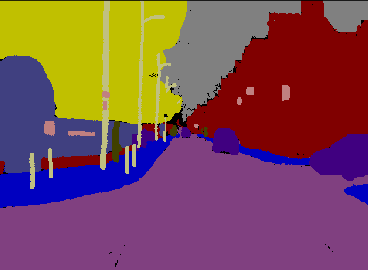} \\ 
\includegraphics[width=26mm]{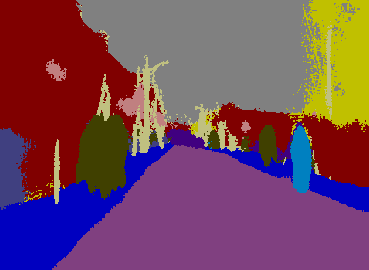} &   \includegraphics[width=26mm]{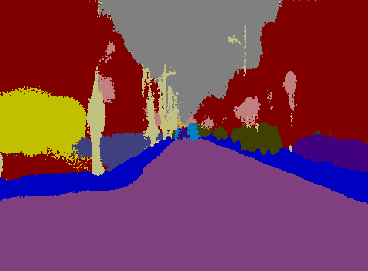} &   \includegraphics[width=26mm]{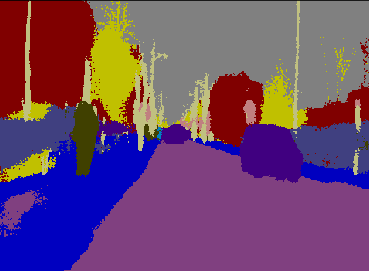} &  
\includegraphics[width=26mm]{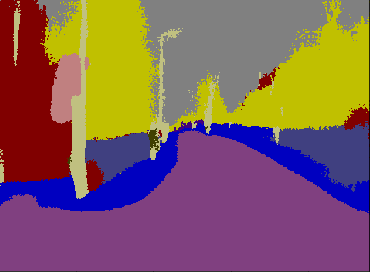} &  
\includegraphics[width=26mm]{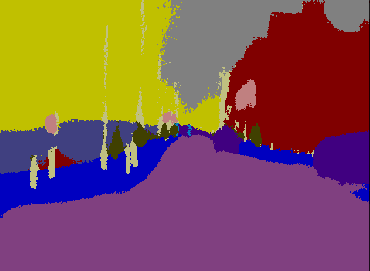} \\ 
\includegraphics[width=26mm]{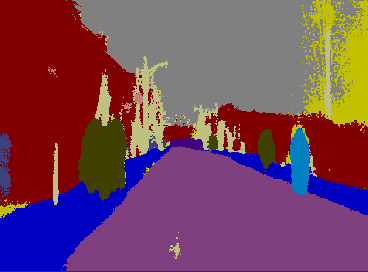} &   \includegraphics[width=26mm]{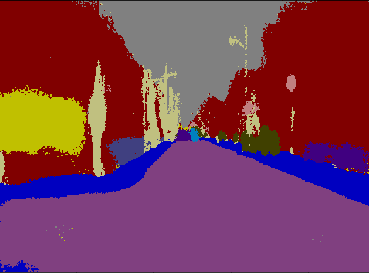} &   \includegraphics[width=26mm]{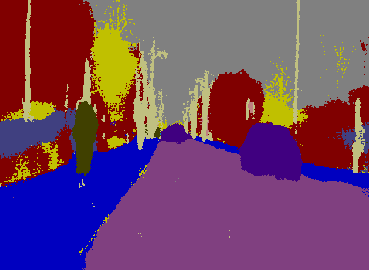} &  
\includegraphics[width=26mm]{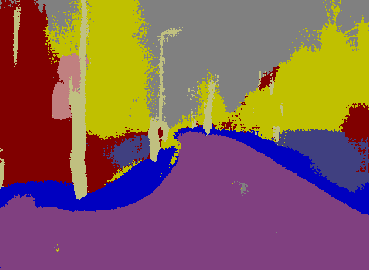} &  
\includegraphics[width=26mm]{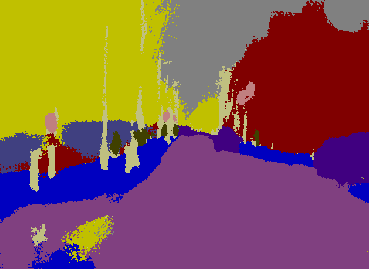} \\ 
\includegraphics[width=26mm]{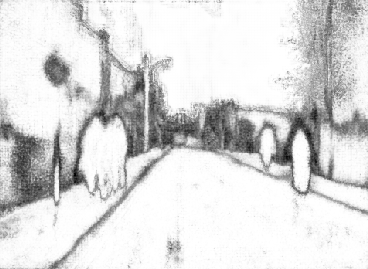} &   \includegraphics[width=26mm]{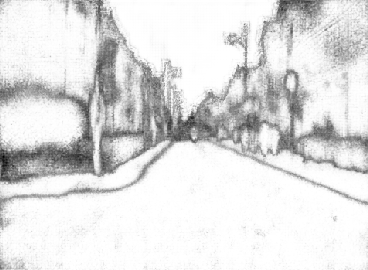} &   \includegraphics[width=26mm]{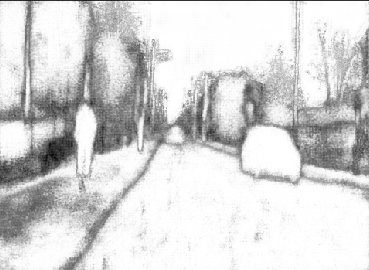} &  
\includegraphics[width=26mm]{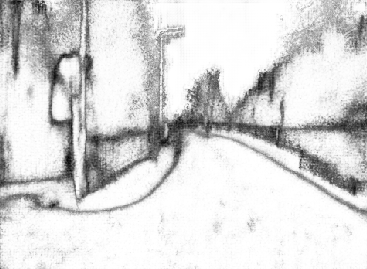} &  
\includegraphics[width=26mm]{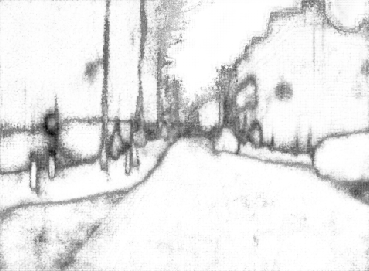} \\
\includegraphics[width=26mm]{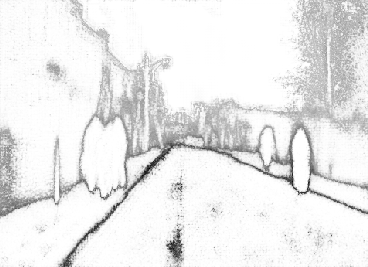} &   \includegraphics[width=26mm]{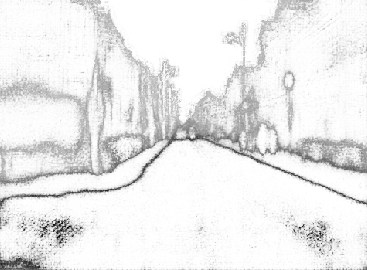} &   \includegraphics[width=26mm]{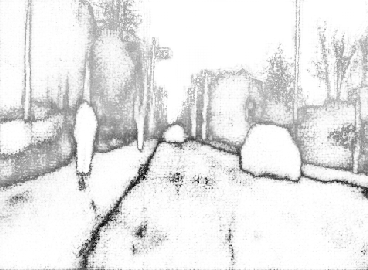} &  
\includegraphics[width=26mm]{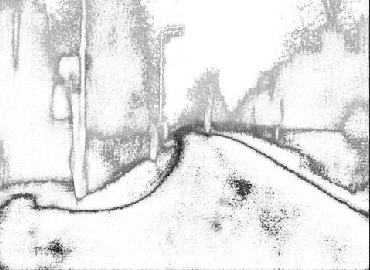} &  
\includegraphics[width=26mm]{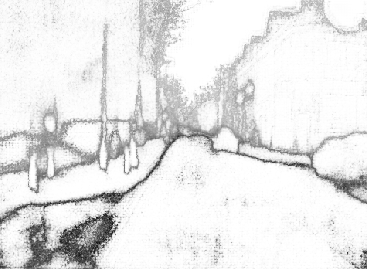} \\
\end{tabular}
\caption{Qualitative results of Bayesian SegNet \cite{Kendall2015BayesianSM} on CamVid \cite{brostow2009semantic}. First row: Original images. Second Row: Ground truth. Third row: Prediction using MC dropout. Fourth row: Our prediction (normal dropout activation scaling). Fifth row: Uncertainty using MC dropout. Sixth row: Our approximation.} \label{qual_encdec}
\end{center}
\end{figure*}

We show more qualitative results of our approximation using Bayesian SegNet\cite{Kendall2015BayesianSM} on CamVid dataset \cite{brostow2009semantic}. These are shown in Fig. \ref{qual_encdec}. We observe that the network is mostly uncertain about object boundaries. We refer to the video in the supplementary material for more qualitative results.

\subsection{Monocular Depth Regression \cite{godard2017unsupervised}}

\begin{figure*}[h]
\begin{center}
\begin{tabular}{c@{\hskip 3pt}c@{\hskip 3pt}c@{\hskip 3pt}c@{\hskip 3pt}c}
\includegraphics[width=26mm]{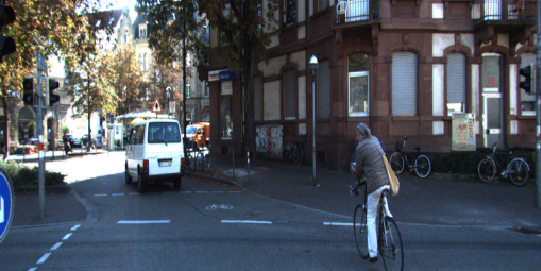}&   \includegraphics[width=26mm]{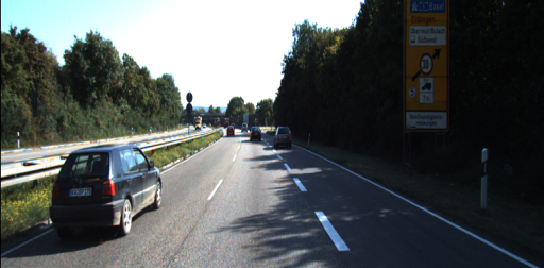}&   \includegraphics[width=26mm]{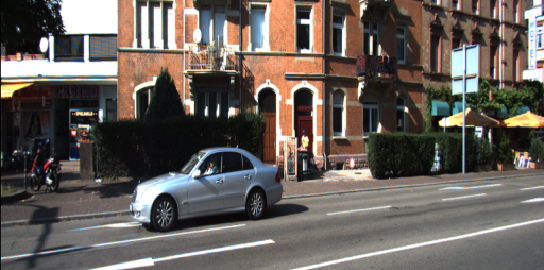}&  \includegraphics[width=26mm]{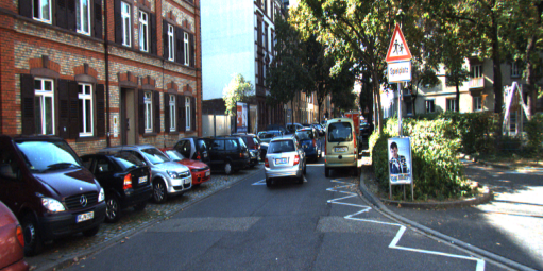}&  \includegraphics[width=26mm]{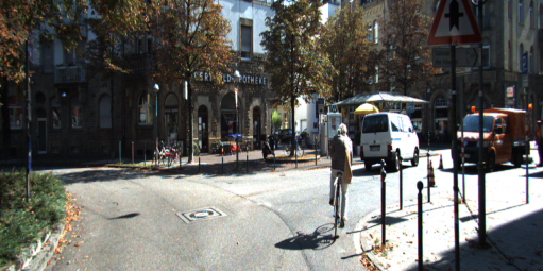}\\ 
\includegraphics[width=26mm]{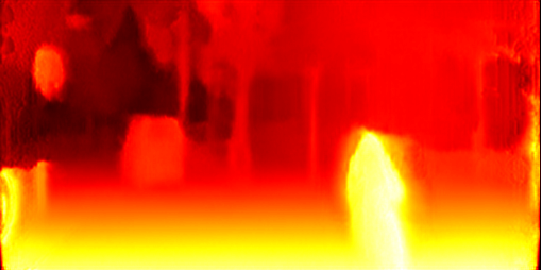} &   \includegraphics[width=26mm]{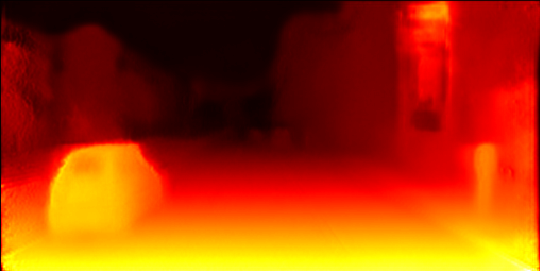} &   \includegraphics[width=26mm]{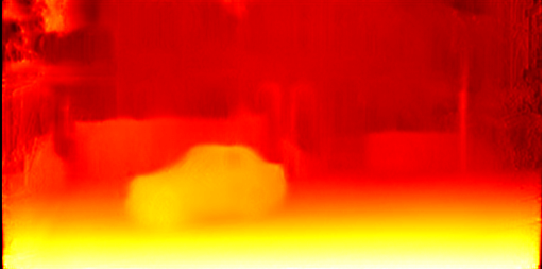} &  
\includegraphics[width=26mm]{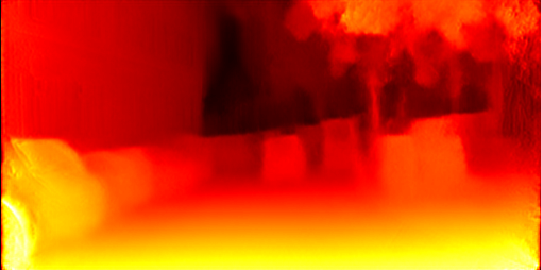} &  
\includegraphics[width=26mm]{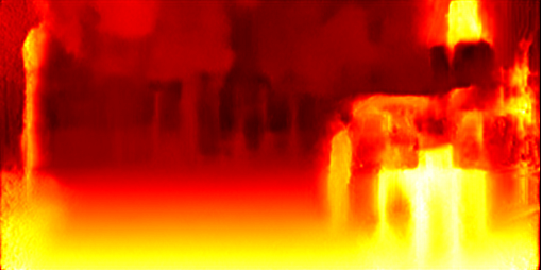} \\ 
\includegraphics[width=26mm]{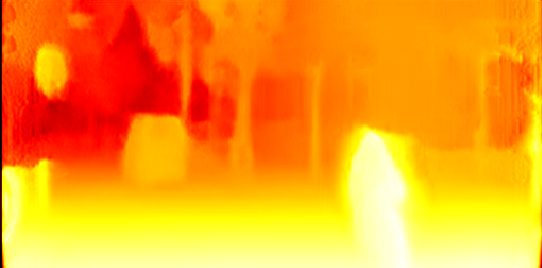}&
\includegraphics[width=26mm]{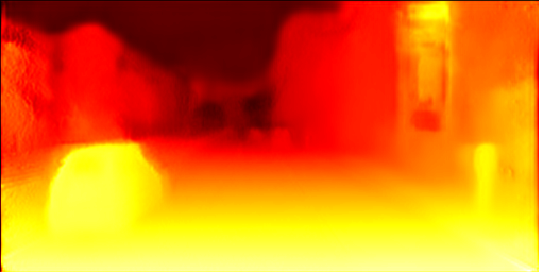} &   \includegraphics[width=26mm]{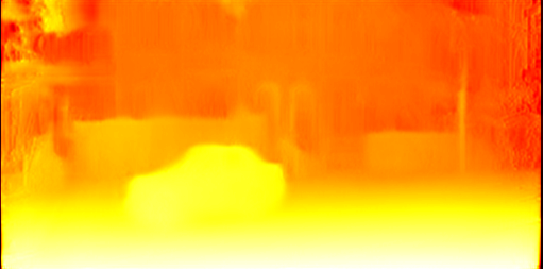} &  
\includegraphics[width=26mm]{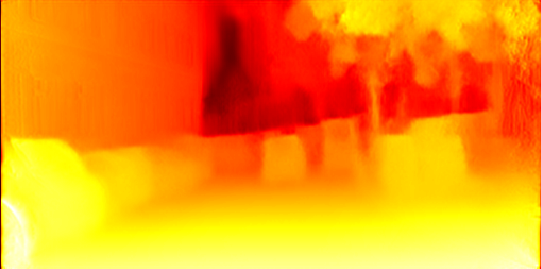} &  
\includegraphics[width=26mm]{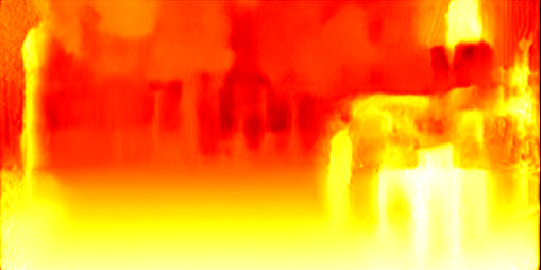} \\ 
\includegraphics[width=26mm]{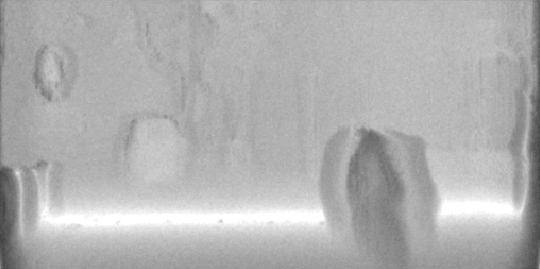} &   \includegraphics[width=26mm]{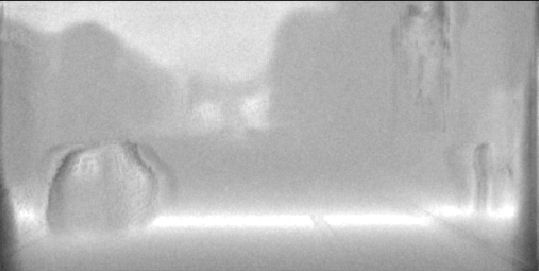} &   \includegraphics[width=26mm]{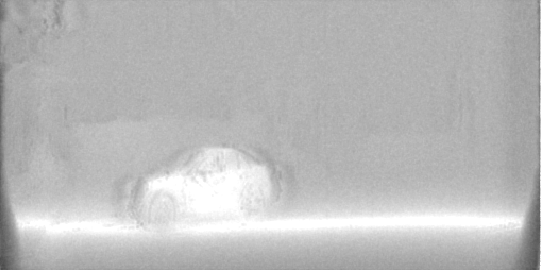} &  
\includegraphics[width=26mm]{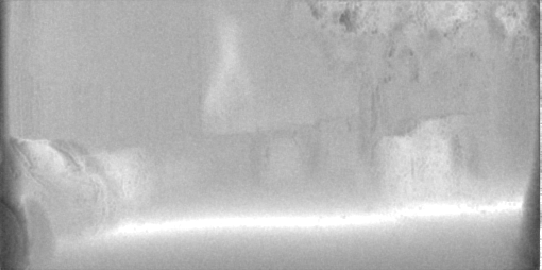} &  
\includegraphics[width=26mm]{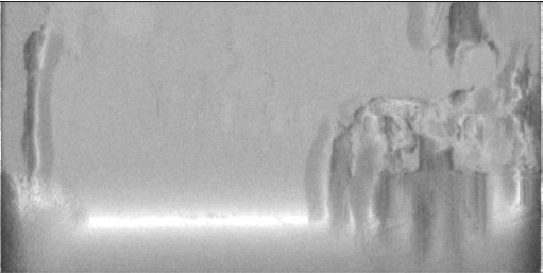} \\ 
\includegraphics[width=26mm]{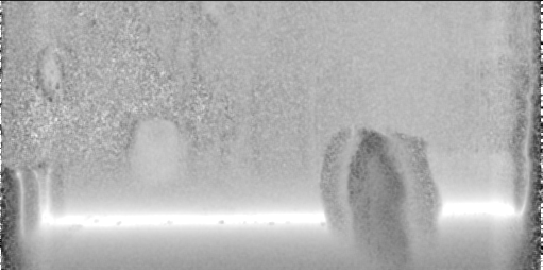} &   \includegraphics[width=26mm]{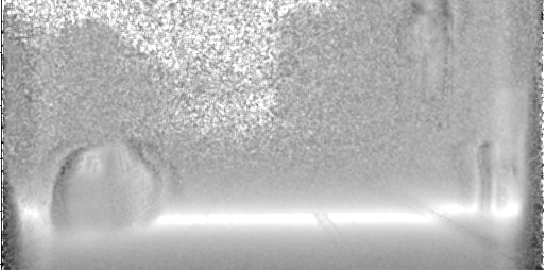} &   \includegraphics[width=26mm]{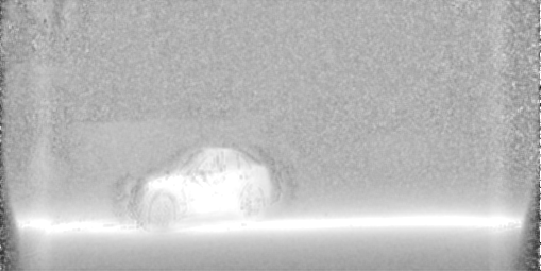} &  
\includegraphics[width=26mm]{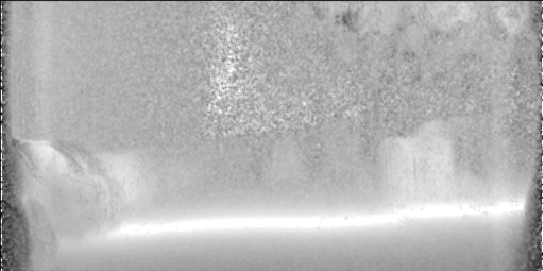} &  
\includegraphics[width=26mm]{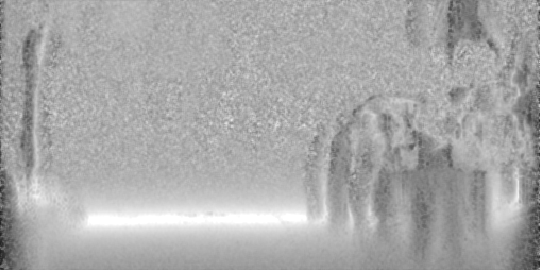} \\
\end{tabular}
\caption{Qualitative results of monocular depth regression \cite{godard2017unsupervised} on KITTI \cite{geiger2013vision}. First row: Original images. Second Row: Prediction using MC dropout. Thrid row: Our prediction (normal dropout activation scaling). Fourth row: Uncertainty using MC dropout. Fifth row: Our approximation.} \label{qual_depth}
\end{center}
\end{figure*}

We show more qualitative results of our approximation using monocular depth regression \cite{godard2017unsupervised} on KITTI dataset \cite{geiger2013vision}. These are shown in Fig. \ref{qual_depth}. We observe that the network is very certain about the region of highest depth resolution and generally uncertain about the left and right border of the image. The latter results from the use of non-overlapping stereo images at training time. We refer to the video in the supplementary material for more qualitative results.

\subsection{Qualitative Results of Class Hold-out}
\begin{figure*}[h]
\centering
\begin{tabular}{c@{\hskip 3pt}c@{\hskip 3pt}c}
  \includegraphics[width=26mm]{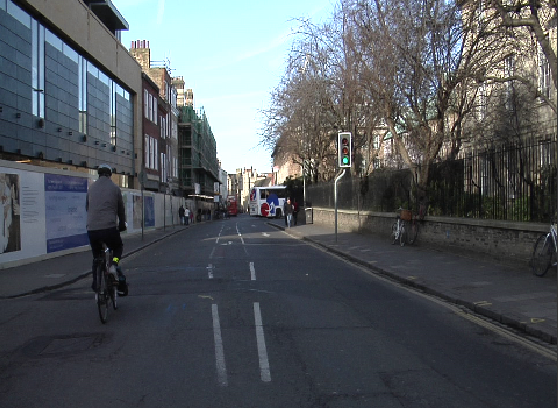} &   \includegraphics[width=26mm]{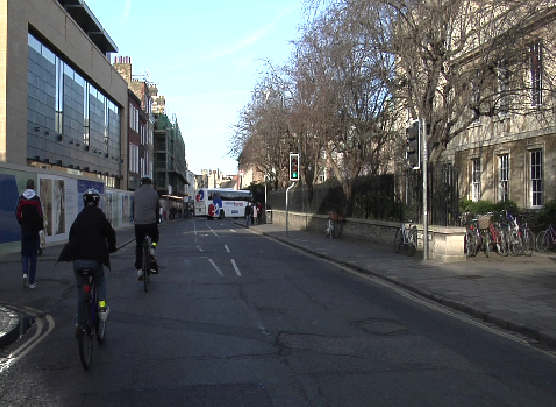} &  \includegraphics[width=26mm]{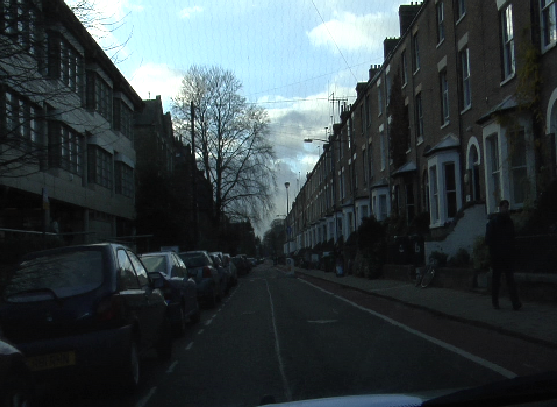}\\ 
  \includegraphics[width=26mm]{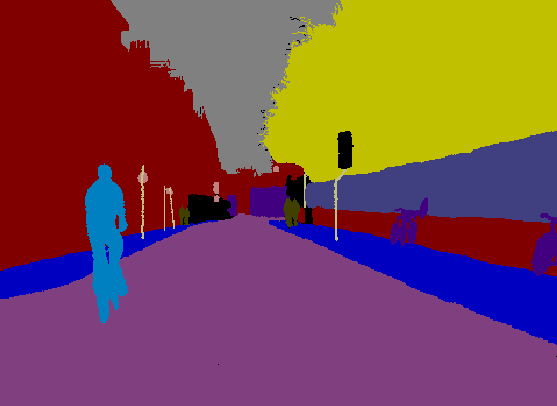} & \includegraphics[width=26mm]{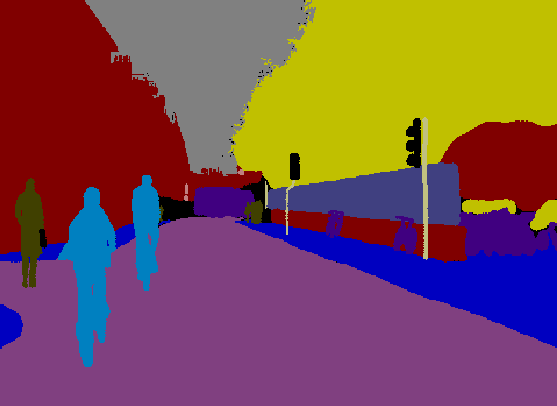} & \includegraphics[width=26mm]{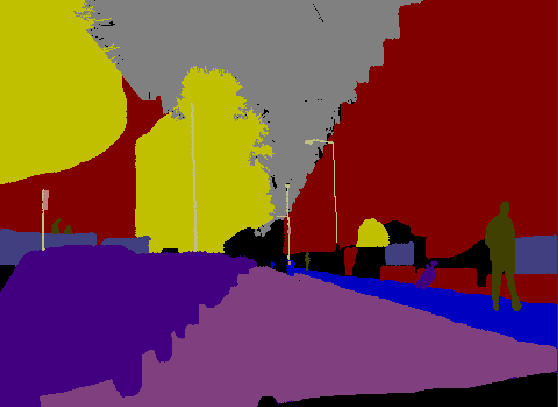}\\
  \includegraphics[width=26mm]{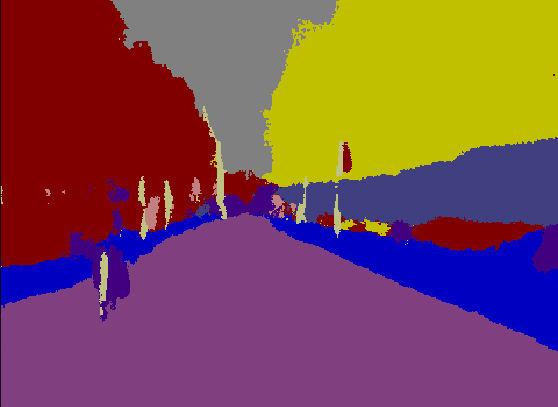} & \includegraphics[width=26mm]{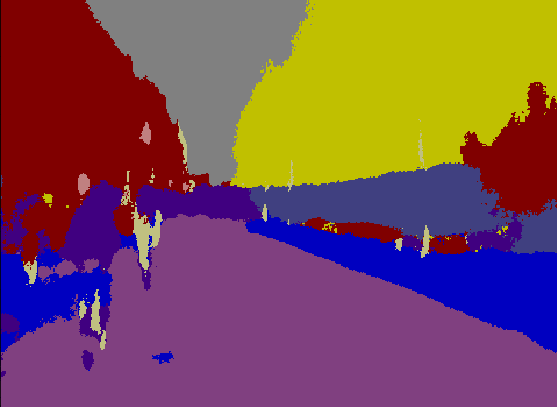} & \includegraphics[width=26mm]{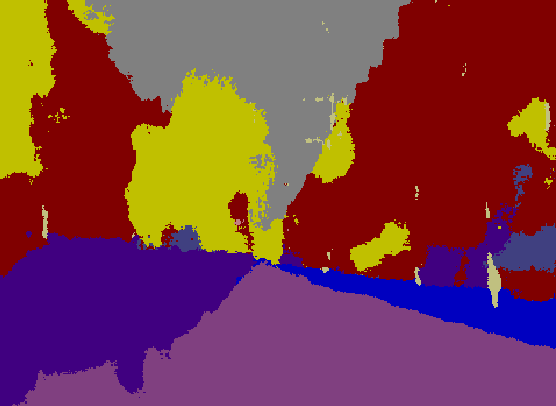}\\
  \includegraphics[width=26mm]{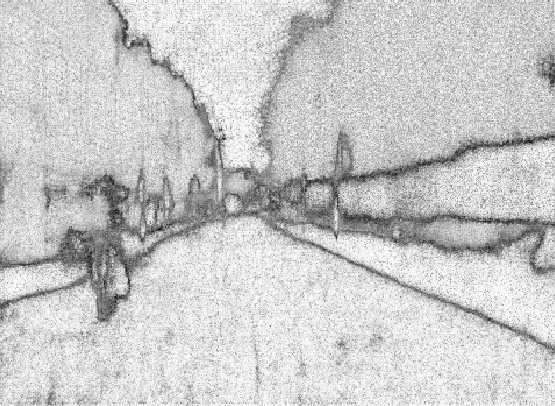} & \includegraphics[width=26mm]{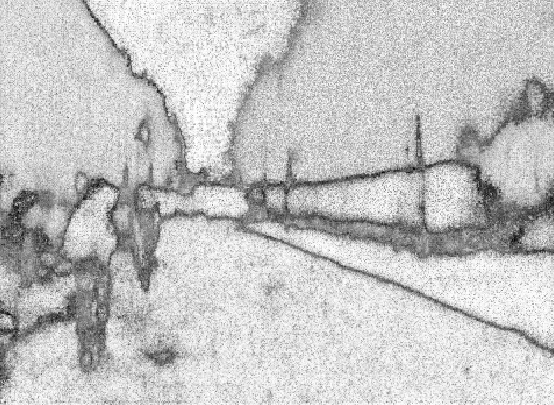} & \includegraphics[width=26mm]{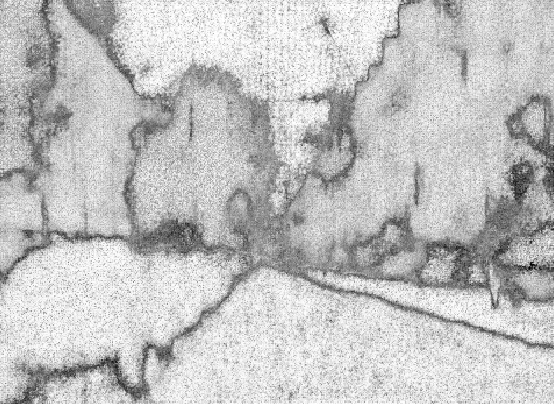}\\
  \includegraphics[width=26mm]{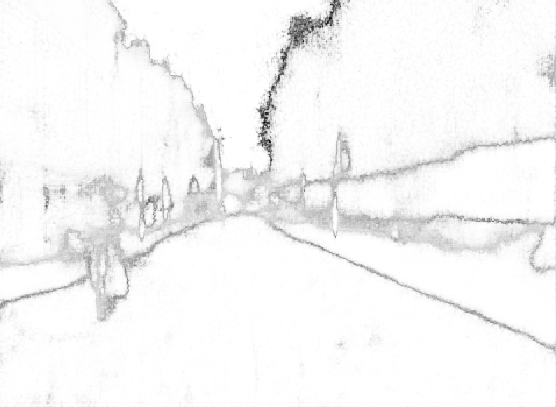} & \includegraphics[width=26mm]{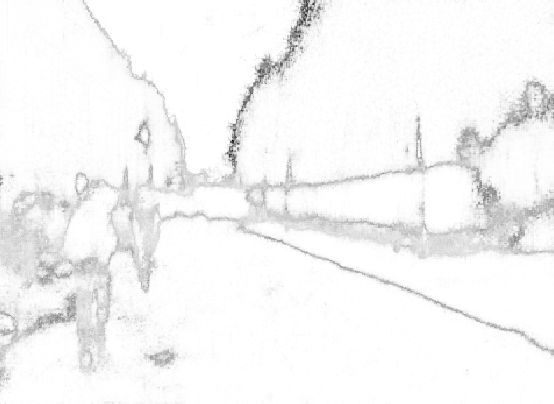} & \includegraphics[width=26mm]{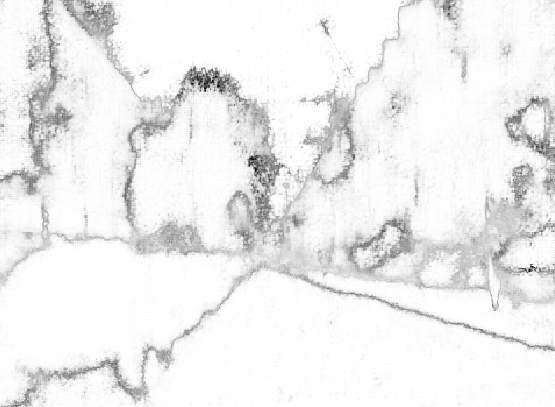}\\
\end{tabular}
\caption{Qualitative results when trained without pedestrian and bicyclist classes. First row: Input image. Second row: Ground truth. Third row: Segmentation result using Monte-Carlo dropout \cite{gal2016dropout}. Fourth row: Uncertainty estimate using Monte-Carlo dropout. Fifth row: Our approximation.}\label{qualitative_bayesian_CLASS_holdout}
\end{figure*}

We trained BayesianSegNet\footnote{only one dropout layer prior to the final layer} withholding the classes for pedestrians and cyclists. Here we show qualitative results of this experiment (see Fig. \ref{qualitative_bayesian_CLASS_holdout}). We observe that the locations of withheld classes tend to be more uncertain than other regions of the image. 

\end{document}